\documentclass[a4paper,fleqn]{cas-sc}
\usepackage{graphicx} 
\usepackage{amsmath, amssymb, amsfonts, amsthm}
\newtheorem{assumption}{Assumption}[section]
\newtheorem{theorem}{Theorem}[section]
\newtheorem{example}{Example}
\usepackage{bm}

\usepackage{mathtools}
\usepackage{algorithm}
\usepackage{float}
\usepackage{algpseudocode}
\usepackage{booktabs}
\usepackage{appendix}
\usepackage[authoryear,longnamesfirst]{natbib}

\date{April 2026}

\def\d{\mbox{d}}
\def\e{\text{e}}
\def\E{\mbox{E}}

\begin{document}
\shorttitle{Sinkhorn Divergence for Reconstructing Random Fields}
\shortauthors{M. Xia and Q. Shen}
\title[mode = title]{A Local Sinkhorn Framework for Conditional Distribution Reconstruction of Multidimensional Random Fields}
\author[1,2]{Mingtao Xia}[type=editor,
                        auid=000,bioid=1,
                        orcid=0000-0002-2116-4712]
\cormark[1]
\ead{xiamingtao97@g.ucla.edu}
\credit{Conceptualization, Methodology, Formal analysis, Investigation, Software, Writing-original draft, Writing-review and editing}

\author[3]{Qijing Shen}[type=editor,
                        auid=000,bioid=1,
                        orcid=0009-0009-6685-0861]
\ead{qijing.shen@ndm.ox.ac.uk}
\credit{Methodology, Formal analysis, Software, Writing-original draft, Writing-review and editing}

\affiliation[1]{organization={School of Mathematics, University of Birmingham},
                addressline={Watson Building, Edgbaston}, 
                city={Birmingham},
                postcode={B15 2TT}, 
                country={United Kingdom}}

\affiliation[2]{organization={Department of Mathematics, University of Houston},
                addressline={3551 Cullen Blvd}, 
                city={Houston},
                postcode={77204}, 
                state={Texas},
                country={United States}}

\affiliation[3]{organization={Nuffield Department of Medicine, University of Oxford, United Kingdom},
                addressline={Old Road Campus}, 
                city={Oxford},
                postcode={OX3 7BN}, 
                country={United Kingdom}}

\cortext[cor1]{Corresponding author}

\numberwithin{equation}{section}
\begin{abstract}
In this paper, we propose a local Sinkhorn divergence framework for conditional distribution reconstruction of multidimensional random fields. By utilizing the debiased Sinkhorn divergence, our proposed approach develops a differentiable and computationally efficient local distribution matching objective to train stochastic neural networks (SNNs). Furthermore, we establish theoretical generalization error estimates for our local Sinkhorn divergence framework, which explicitly characterizes the trade-off between approximation bias and statistical efficiency controlled by the regularization parameter and reveals how our proposed local Sinkhorn divergence loss function can be efficiently applied to learning multidimensional random field models. The proposed framework provides a scalable alternative to exact local optimal transport for conditional distribution reconstruction, offering a practical compromise between geometric fidelity, statistical efficiency, and computational scalability for uncertainty quantification and probabilistic scientific machine learning.
Through various numerical examples, we compare our proposed local Sinkhorn divergence framework with other loss functions to train SNNs and with other machine-learning-based uncertainty quantification frameworks, demonstrating that the proposed local Sinkhorn divergence framework achieves an effective balance between reconstruction accuracy and computational efficiency while maintaining good scalability for multidimensional stochastic systems.

\end{abstract}

\begin{highlights}
\item  We propose a scalable local entropically regularized optimal transport framework to train stochastic neural networks for learning random fields.
\item We derive generalization error bounds that reveal how our proposed framework could efficiently learn multidimensional random fields.
\item Our proposed approach accurately reconstructs stochastic random fields and dynamical systems and outperforms several machine-learning-based benchmark approaches in uncertainty quantification.
\end{highlights}

\maketitle

\section{Introduction}
The identification of random fields from observational data has become an increasingly important topic in uncertainty quantification, stochastic modeling, and scientific machine learning. Many physical and engineering systems, including porous media flow, turbulent transport, groundwater dynamics, and biological processes, are inherently affected by uncertain parameters or random forcing, leading to stochastic responses that cannot be adequately characterized by deterministic models alone \cite{ghanem1991stochastic,xiu2010numerical,sullivan2015introduction}. Consequently, instead of predicting only the conditional expectation, it is often desirable to reconstruct the entire conditional probability distribution of the stochastic solution, thereby providing a more comprehensive description of uncertainty propagation. However, the primary challenge is usually not learning the conditional mean, but accurately reconstructing the multidimensional conditional distributions in a computationally scalable manner.

Recent advances in machine learning have enabled data-driven approximations of multidimensional stochastic systems without requiring explicit knowledge of the underlying governing equations. Various deep generative models, including conditional variational autoencoders (CVAEs), generative adversarial networks (GANs), conditional normalizing flows (CNFs), and diffusion models, have demonstrated remarkable capability in learning complex probability distributions from data \cite{kingma2013auto,goodfellow2014gan,rezende2015variational,ho2020denoising}. 
Nevertheless, for stochastic random field reconstruction, the learned distributions should preserve the geometric structure of probability measures associated with nearby physical states. This requirement naturally motivates the use of optimal transport (OT), whose Wasserstein metric provides a physically meaningful distance between probability distributions and has recently attracted increasing attention in scientific machine learning \cite{villani2009optimal,peyre2019computational}. Compared with divergence-based discrepancies, such as the Kullback--Leibler divergence or maximum mean discrepancy \cite{gretton2012kernel}, Wasserstein distances remain informative even when two probability measures have disjoint supports and provide meaningful gradients during optimization \cite{arjovsky2017wasserstein,peyre2019computational}. These advantages have motivated numerous OT-based machine learning methods, including Wasserstein GANs, gradient-flow formulations, and OT regularization for scientific computing \cite{frogner2015learning,arjovsky2017wasserstein}. However, the exact computation of Wasserstein distances requires solving a large-scale linear programming problem whose complexity grows rapidly with sample size, making repeated evaluations during neural network training prohibitively expensive. Our recent work introduced a local squared Wasserstein-$2$ ($W_2$) framework for random field reconstruction, in which neighboring samples associated with nearby input states were compared through local optimal transport distances. 
By exploiting the locality of conditional distributions, the method was shown to accurately reconstruct stochastic fields from sparse observations while preserving important geometric structures of the underlying probability measures \cite{xia2026local}. However, the local squared $W_2$ distance relies on repeatedly solving exact optimal transport problems during neural network training. As the number of neighborhood evaluations and samples increases, the computational cost becomes the dominant bottleneck, significantly limiting scalability for multidimensional stochastic systems and large datasets. Additionally, the generalization error bound may deteriorate when the squared $W_2$ distance between two probability measures is estimated using empirical probability measures, particularly in high-dimensional settings.

Recently, Cuturi introduced entropic regularization of optimal transport, leading to the Sinkhorn algorithm that dramatically accelerates OT computations while remaining fully differentiable \cite{cuturi2013sinkhorn}. Subsequently, Feydy \textit{et al.} proposed the Sinkhorn divergence, which removes the entropic bias and interpolates smoothly between the exact Wasserstein distance and kernel-based maximum mean discrepancy \cite{feydy2019interpolating}. The Sinkhorn divergence has since been successfully applied to generative modeling, Bayesian inference, and scientific machine learning because of its computational efficiency and stable gradient properties \cite{genevay2018learning,feydy2019interpolating}. Furthermore, through introducing an extra entropic regularization term, the generalization error when estimating the Sinkhorn divergence using the empirical measures may not deteriorate with the dimensionality of the associated random variables \cite{genevay2018learning}.

Motivated by these developments, this work develops a local Sinkhorn-divergence framework for conditional distribution reconstruction of stochastic fields. Building upon our previous local $W_2$ formulation, the proposed approach replaces the expensive exact OT computation with Sinkhorn divergence, resulting in substantially improved computational efficiency while preserving the local geometric information of conditional probability distributions. The proposed local Sinkhorn divergence bridges exact optimal transport and kernel-based distribution matching by introducing an entropic regularization that provides a controllable bias while significantly improving computational efficiency and statistical scalability. Owing to the differentiability and favorable numerical stability of Sinkhorn divergence, the resulting learning framework is considerably more efficient in reconstructing conditional distributions of the to-be-learned random fields. Furthermore, we shall show from both theoretical and empirical aspects that incorporating the regularization term could lead to improved accuracy when reconstructing multidimensional random field models using empirical distributions. 
Numerical experiments on stochastic Darcy flow and multidimensional stochastic FitzHugh--Nagumo (FHN) systems demonstrate that the proposed approach achieves reconstruction accuracy comparable to that of the exact Wasserstein formulation while significantly reducing the overall training cost.

The main contributions of this work are summarized as follows.

\begin{itemize}
\item
We extend our recent local optimal transport framework for random field reconstruction from the exact Wasserstein distance to the debiased Sinkhorn divergence, leading to a computationally efficient, scalable, and fully differentiable local distribution matching method for training SNNs to reconstruct random fields.
\item
We provide a theoretical analysis of the proposed local Sinkhorn formulation by combining the neighborhood approximation error with the regularization and statistical convergence properties of empirical Sinkhorn divergence. The resulting generalization bounds reveal how the local Sinkhorn divergence may partially alleviate the curse of dimensionality through introducing an entropic regularization term.
\item
Through various numerical examples, we demonstrate that the proposed local Sinkhorn divergence provides an effective compromise between approximation accuracy and computational efficiency and outperforms several machine-learning-based uncertainty quantification benchmarks, making it particularly suitable for multidimensional uncertainty quantification problems.
\end{itemize}

The remainder of this paper is organized as follows. Section~\ref{section2} introduces the proposed local Sinkhorn divergence framework together with its theoretical analysis and generalization error bounds. Section~\ref{section3} presents three numerical examples, including one-dimensional conditional distribution reconstruction, stochastic Darcy flow, and stochastic FHN systems, where the proposed method is compared with several local distribution matching losses and existing machine-learning benchmark approaches. Finally, Section~\ref{section4} concludes the paper with a summary of the main findings and discusses several directions for future research. Frequently used notations throughout this paper are summarized in Table~\ref{tab:notation}.

\begin{table}[t]
\centering
\caption{Frequently used notations throughout this paper.}
\label{tab:notation}
\renewcommand{\arraystretch}{1.15}
\begin{tabular}{ll}
\toprule
Notation & Description\\
\midrule
$\bm{x}\in\mathcal D$ &
Input variable.
\\
$\mathcal D\subseteq\mathbb{R}^n$ &
Input domain.
\\
$N$ &
Total number of training samples.
\\
$\omega$ ($\hat\omega$) &
Random variable representing uncertainty.
\\

$\bm{y}(\bm x,\omega)\in\mathbb{R}^d$ &
Target random field.
\\

$\hat{\bm{y}}(\bm{x},\hat\omega)\in\mathbb{R}^d$ &
Reconstructed random field.
\\

$\mu_{\bm{x}}$ &
Conditional probability distribution of $\bm{y}(\bm{x},\omega)$.
\\

$\hat\mu_{\bm{x}}$ &
Conditional probability distribution of $\hat{\bm{y}}(\bm{x},\hat \omega)$.
\\

$\mu_{\bm{x},\delta}^{\e}$ &
Empirical $\delta$-neighborhood measure associated with $\bm{y}(\bm{x},\omega)$ (details given in Section~\ref{section2}).
\\

$\hat\mu_{\bm{x},\delta}^{\e}$ &
Empirical $\delta$-neighborhood measure associated with $\hat{\bm{y}}(\bm{x},\hat{\omega})$ (details given in Section~\ref{section2}).
\\


$\delta$ &
Neighborhood radius.
\\



$\nu$ &
Probability measure on the input domain.
\\

$\nu^{\e}$ &
Empirical measure induced by the observed inputs.
\\

$W_2^2(\mu,\hat\mu)$ &
Squared $W_2$ distance between the two probability measures $\mu, \hat\mu$.
\\

$S_\varepsilon(\mu,\hat\mu)$ &
Debiased Sinkhorn divergence between the two probability measures $\mu, \hat\mu$.
\\

$\varepsilon$ &
Sinkhorn regularization parameter.
\\

$\overline{\mathcal W}_2$ &
Averaged squared Wasserstein discrepancy.
\\


$\overline{\mathcal S}_{\varepsilon,\delta}^{\e}$ &
Averaged empirical local Sinkhorn divergence.

\\





\bottomrule
\end{tabular}
\end{table}

\section{A local Sinkhorn divergence method for conditional distribution reconstruction of random fields}
\label{section2}
In this section, we formally introduce the random field model we reconstruct and the local Sinkhorn divergence framework we propose. 
Consider the reconstruction of conditional probability distributions from scattered observations. Let
\begin{equation}
\bm{y}_{\bm{x}}=f(\bm{x},\omega),
\qquad
x\in\mathcal D\subset\mathbb R^n,
\label{eq:true_model}
\end{equation}
be the unknown random field to be reconstructed, where $\omega\in\Omega$ denotes the underlying random variable and $f:\mathcal D\times\Omega\rightarrow\mathbb R^d$ is an unknown stochastic mapping. For each input location $\bm{x}$, only realizations of the random response $y(\bm{x},\omega)$ are available, while the conditional probability measure of $\bm{y}(\bm{x};\omega)$, denoted as $\mu_{\bm{x}}$,
is unknown.

Our objective is to construct an approximation
\begin{equation}
\hat{\bm{y}}_{\bm{x}}
=
\hat f(\bm{x},\hat{\omega}),
\qquad
\bm{x}\in\mathcal D,
\label{eq:model}
\end{equation}
where
$\hat f$ is another stochastic mapping, and $\hat{\omega}\in\hat{\Omega}$ is another random variable. The goal is to match the distribution of $\bm{y}(\bm{x};\omega)$ using the distribution of $\hat{\bm{y}}(\bm{x};\hat{\omega})$, denoted by $\hat{\mu}_{\bm{x}}$ for every $\bm{x}\in\mathcal D$.

\begin{figure}[h]
\centering
\includegraphics[width=.85\textwidth]{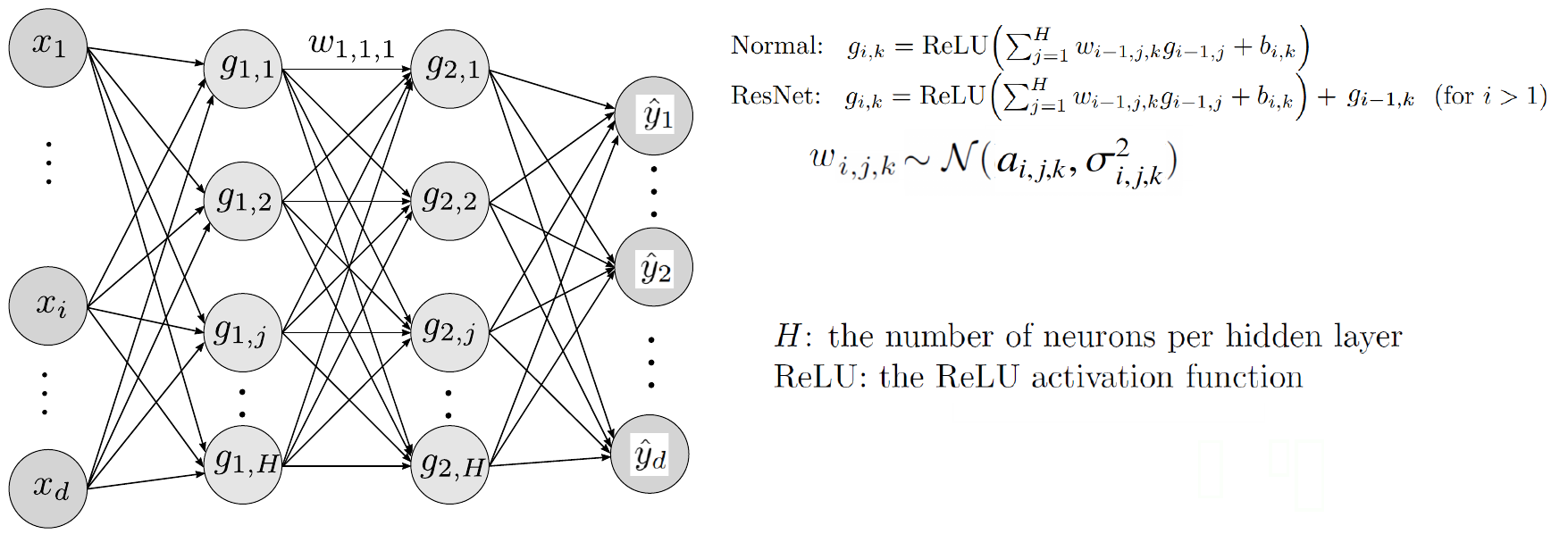}
\caption{
Illustration of the SNN used for conditional distribution reconstruction.
Each linear layer samples its weights from Gaussian distributions whose means and variances are trainable during forward propagation, enabling stochastic realizations to be generated for the same input.
Residual connections (ResNet) could be employed to improve optimization stability.
The generated samples are used to construct empirical conditional distributions of the reconstructed $\hat{\bm{y}}(\bm{x};\hat{\omega})$ in Eq.~\eqref{eq:model}. The stochastic weights in the SNN serve as the random variable $\hat{\omega}$ in Eq.~\eqref{eq:model}. The \texttt{ReLU} activation function may be replaced with other activation functions.
}
\label{fig:snn}
\end{figure}

We employ a stochastic neural network (SNN) similar to our previous work \cite{xia2026local} as the approximate random field model $\hat{f}$ in Eq.~\eqref{eq:model}. The structure of the SNN is described in Fig.~\ref{fig:snn}.
Unlike deterministic neural networks, the SNN contains random parameters (as the random variable $\hat{\omega}$ in Eq.~\eqref{eq:model}) sampled during forward propagation, enabling the model to generate multiple realizations for the same input location. Furthermore, it is proved in \cite[Appendix H]{xia2025new} that this SNN can approximate any random field model under an averaged squared $W_2$ metric, subject to nonrestrictive assumptions.


First, we briefly review the Sinkhorn divergence between two probability measures. Let $\bm{y},\hat{\bm{y}}\in\mathbb{R}^{n}$ be two random variables satisfying:
\begin{equation}
\mathbb{E}\!\left[\|\bm{y}\|^{2}\right]<\infty,
\qquad
\mathbb{E}\!\left[\|\hat{\bm{y}}\|^{2}\right]<\infty .
\label{eq:moment}
\end{equation}
Throughout this paper, $\|\cdot\|$ denotes the $\ell^2$ norm of a vector. The probability distributions of $\bm{y}$ and $\hat{\bm{y}}$ are denoted by $\mu$ and $\hat\mu$, respectively.
Let
\begin{equation}
    c(y,\hat y)=\|\bm{y}-\hat{\bm{y}}\|^{2}
    \label{quadratic_cost}
\end{equation}
be the quadratic transportation cost.
The entropically regularized optimal transport cost is defined as:
\begin{equation}
W_{\varepsilon}^2(\mu,\hat\mu)
=
\min_{\pi\in\Pi(\mu,\hat\mu)}
\left\{
\int_{\mathbb{R}^{n}\times\mathbb{R}^{n}}
c(\bm{y},\hat{\bm{y}})\,
\d\pi(\bm{y},\hat{\bm{y}})
+
\varepsilon
\,
\mathrm{KL}
(\pi\,\|\,\mu\otimes\hat\mu)
\right\},
\label{eq:sinkhornOT}
\end{equation}
where $\Pi(\mu,\hat\mu)$ denotes the set of probability measures on
$\mathbb{R}^{n}\times\mathbb{R}^{n}$
whose marginals are $\mu$ and $\hat\mu$, respectively. 
In Eq.~\eqref{eq:sinkhornOT},
\[
\mathrm{KL}(\pi\,\|\,\mu\otimes\hat\mu)
=
\int_{\mathbb{R}^{n}\times\mathbb{R}^{n}}
\log\!\left(
\frac{\d\pi}{\d(\mu\otimes\hat\mu)}
\right)
\,\d\pi
\]
denotes the Kullback--Leibler divergence of $\pi$ with respect to the product measure
$\mu\otimes\hat\mu$, provided that
$\pi\ll\mu\otimes\hat\mu$; otherwise,
$\mathrm{KL}(\pi\,\|\,\mu\otimes\hat\mu)=+\infty$, and 
$\varepsilon>0$
is the entropic regularization parameter.

Unlike the classical Wasserstein distance, the regularized transport cost does not vanish when
$\mu=\hat\mu$.
To remove this entropic bias, the Sinkhorn divergence is defined as
\begin{equation}
S_{\varepsilon}(\mu,\hat\mu)
=
W_{\varepsilon}^2(\mu,\hat\mu)
-
\frac12
W_{\varepsilon}^2(\mu,\mu)
-
\frac12
W_{\varepsilon}^2(\hat\mu,\hat\mu).
\label{eq:sinkhorndiv}
\end{equation}
The Sinkhorn divergence in Eq.~\eqref{eq:sinkhorndiv} is symmetric, non-negative, differentiable with respect to the empirical samples, and satisfies
$S_{\varepsilon}(\mu,\mu)=0$; furthermore, the Sinkhorn divergence converges to the Wasserstein distance as $\varepsilon\rightarrow0$ while remaining differentiable and computationally tractable, whereas choosing $\varepsilon\rightarrow\infty$ recovers the maximum mean discrepancy \cite{feydy2019interpolating}. Yet, the Sinkhorn divergence is more computationally efficient through introducing an additional regularizing term in Eq.~\eqref{eq:sinkhornOT} while preserving the advantage of the squared $W_2$ distance for comparing two probability measures. 

To simplify the subsequent analysis, we introduce the following assumptions.

\begin{assumption}
\rm
\label{assumptions}
Assume that the uncertainty model
$\bm{y}_{\bm{x}}$
and the approximation model $\hat{\bm{y}}_{\bm{x}}$ in Eqs.~\eqref{eq:true_model} and \eqref{eq:model} satisfy the following conditions.

\begin{enumerate}

\item
Both random fields are uniformly bounded, \textit{i.e.}, there exists
$M_0>0$
such that

\begin{equation}
\|\bm{y}(\bm{x},\omega)\|
\le
M_0,
\qquad
\|\hat{\bm{y}}(\bm{x},\hat\omega)\|
\le
M_0,
\qquad
\forall \bm{x}\in \mathcal{D}.
\label{eq:bounded}
\end{equation}

\item
The mappings
$f$
and
$\hat f$
are uniformly Lipschitz continuous with respect to the input variable.
That is, there exists a constant
$L>0$
such that

\begin{equation}
\|f(\bm{x},\omega)-f(\bm{x}',\omega)\|
\le
L\|\bm{x}-\bm{x}'\|, \,\,\,\,
\|\hat f(\bm{x},\hat\omega)-\hat f(\bm{x}',\hat\omega)\|
\le
L\|\bm{x}-\bm{x}'\|, \,\, \forall\bm{x},\bm{x}'\in \mathcal{D}.
\label{eq:lipschitz}
\end{equation}


\item

The random variables
$\omega$
and
$\hat\omega$
are independent of the input variable
$\bm{x}$.

\end{enumerate}
\end{assumption}

Following our previous work on local optimal transport \cite{xia2026local}, we define the weighted average squared $W_2$ metric between the ground-truth and approximate random fields $\bm{y}_{\bm{x}}$ and $\hat{\bm{y}}_{\bm{x}}$ in Eqs.~\eqref{eq:true_model} and \eqref{eq:model}:
\begin{equation}
\overline{\mathcal{W}}_{2}(\bm{y}_{\bm{x}},\hat{\bm{y}}_{\bm{x}})
\coloneqq
\int_{\mathcal{D}}
W_2^2
(\mu_{\bm{x}},\hat\mu_{\bm{x}})
\,\nu(\d \bm{x}).
\label{eq:globalsinkhorn}
\end{equation}



$\overline{\mathcal{W}}_{2}(\bm{y}_{\bm{x}},\hat{\bm{y}}_{\bm{x}})$ defined in Eq.~\eqref{eq:globalsinkhorn} is nonnegative, and 
if the weighting measure of the input variable
$\nu$
is strictly positive over
$\mathcal{D}$,
then 
$\overline{\mathcal{W}}_{2}(\bm{y}_{\bm{x}},\hat{\bm{y}}_{\bm{x}})=0$
holds if and only if
\[
W_2^2(\mu_{\bm{x}},\hat\mu_{\bm{x}})=0,
\qquad
\text{for almost every }
\bm{x}\in \mathcal{D},
\]
which implies that the probability measure of $\bm{y}_{\bm{x}}$ can be matched by the probability measure of $\hat{\bm{y}}_{\bm{x}}$ for almost every $\bm{x}\in\mathcal{D}$.
Consequently, minimizing
(\ref{eq:globalsinkhorn})
is equivalent to matching the conditional distributions $\mu_{\bm{x}}$ using $\hat{\mu}_{\bm{x}}$ throughout the computational domain.
However, in practical applications only a finite set of scattered observations
$\{
(\bm{x}_i,\bm{y}_{\bm{x}_i})
\}_{i=1}^{N}$
is available.
Since the conditional probability measures
$\mu_{\bm{x}}$
and
$\hat\mu_{\bm{x}}$
cannot be evaluated directly from finite observations, we approximate them by local empirical measures constructed from neighboring samples.

Given a training sample $\bm{x}_i\in\{\bm{x}_1,...\bm{x}_N\}$, we
define its neighborhood
\begin{equation}
B(\bm{x}_i,\delta)
=
\{
\bm{x}_j:
\|\bm{x}_j-\bm{x}_i\|<\delta,\,\, j=1,...,N
\},
\label{eq:neighbor}
\end{equation}
where $\delta>0$ is the neighborhood radius. For implementing the neighborhood technique, we randomly choose a subset 
\begin{equation}
\mathcal{B}(\bm{x}_i, \delta)\subseteq B(\bm{x}_i, \delta).
    \label{tilde_B_def}
\end{equation}
Let $\mu_{x_i,\delta}^{\e}$ and $\hat\mu_{\bm{x}_i,\delta}^{\e}$ denote the corresponding empirical conditional probability measures constructed from the neighboring observations, \textit{i.e.},
$\mu_{\bm{x}_i,\delta}^{\e}$ is the empirical distribution of $\bm{y}(\bm{x};\omega), \bm{x}\in \mathcal{B}(\bm{x}_i, \delta)$ and $\hat\mu_{\bm{x}_i,\delta}^{\e}$ is the empirical distribution of $\hat{\bm{y}}(\bm{x};\omega), \bm{x}\in \mathcal{B}(\bm{x}_i, \delta)$. $\mu_{\bm{x}_i,\delta}^{\e}$ and $\hat\mu_{\bm{x}_i,\delta}^{\e}$ are used to approximate $\mu_{\bm{x}}$ and $\hat{\mu}_{\bm{x}}$, respectively.
We then introduce the proposed local Sinkhorn divergence loss:
\begin{equation}
S_{\varepsilon,\delta}^{\e}(\bm{y}_{\bm{x}},\hat{\bm{y}}_{\bm{x}})
\coloneqq
\int_{\mathcal{D}}
S_{\varepsilon}
(
\mu_{\bm{x},\delta}^{\e},
\hat\mu_{\bm{x},\delta}^{\e}
)
\,\nu^{\e}(\d\bm{x}),
\label{eq:localsinkhorn}
\end{equation}
where
$\nu^{\e}$
denotes the empirical measure associated with the observed input variable $\bm{x}$.
Eq.~\eqref{eq:localsinkhorn}
is not a single Sinkhorn divergence between two probability measures.
Instead, it computes the average Sinkhorn divergence between the local empirical conditional distributions over the entire input domain.
The superscript
$\e$
indicates empirical probability measures estimated from neighboring observations, while the subscript
$\delta$
denotes the neighborhood radius.


Compared with the $W_2$ metrics, the Sinkhorn divergence introduces an entropic regularization, allowing the optimal transport problem to be solved efficiently by iterative matrix scaling rather than network simplex optimization \cite{cuturi2013sinkhorn,feydy2019interpolating}. Consequently, the resulting loss is fully differentiable and naturally amenable to GPU-parallel computation, while recent theoretical results also demonstrate improved statistical convergence properties for empirical Sinkhorn divergences compared with empirical Wasserstein distances \cite{genevay2019sample}. Specifically, we establish two different forms of the generalization error bounds on how the local Sinkhorn divergence Eq.~\eqref{eq:localsinkhorn} approximates the averaged squared Wasserstein discrepancy in Eq.~\eqref{eq:globalsinkhorn}.

\begin{theorem}[Generalization bound based on the generalization error of the local squared $W_2$ loss]
\rm
\label{thm:sinkhorn_neighborhood_generalization}
Under Assumptions~\ref{assumptions}, let $\mu_{\bm{x}}$ and $\hat\mu_{\bm{x}}$ denote the true conditional probability measures associated with $\bm{y}_{\bm{x}}$ and $\hat{\bm{y}}_{\bm{x}}$ at location $\bm{x}\in\mathcal D$, respectively. 
Let $\mu_{\bm{x},\delta}^{\e}$ and $\hat\mu_{\bm{x},\delta}^{\e}$ be the corresponding empirical neighborhood measures built from the sample set $\{(\bm{x}_i,\bm{y}_{\bm{x}_i})\}_{i=1}^{N}$ with neighborhood radius $\delta>0$. We assume $\varepsilon<\frac{2M_0^2}{\sqrt{d}}$. For $\overline{\mathcal S}_{\varepsilon,\delta}^{\e}(\bm{y}_{\bm{x}},\hat{\bm{y}}_{\bm{x}})$ and $\overline{\mathcal W}_{2}(\bm{y}_{\bm{x}},\hat{\bm{y}}_{\bm{x}})$ defined in Eqs.~\eqref{eq:globalsinkhorn} and \eqref{eq:localsinkhorn}, 
there exists a constant $C>0$ such that:
\begin{equation}
\begin{aligned}
\mathbb E\!\left[
\left|
\overline{\mathcal S}_{\varepsilon,\delta}^{\e}(\bm{y}_{\bm{x}},\hat{\bm{y}}_{\bm{x}})
-
\overline{\mathcal W}_{2}(y_x,\hat y_x)
\right|
\right]
\le
2\varepsilon\log\!\left(
\frac{2e^2M_0^2}{\sqrt d\,\varepsilon}
\right)
+
\frac{4M_0}{\sqrt{N}}
+
8CM_0\,\mathbb E\!\bigl[h(N(x,\delta),d)\bigr]
+16M_0^2\,\delta,
\end{aligned}
\label{eq:sinkhorn_gen_bound}
\end{equation}
where $N$ is the number of training samples, and 
\begin{equation}
    N(\bm{x},\delta) \coloneqq |\mathcal{B}(\bm{x}, \delta)|
    \label{N_neighborhood}
\end{equation}
denotes the effective number of samples chosen from the neighborhood $B(\bm{x}, \delta)$ centered at $\bm{x}$ defined in Eq.~\eqref{eq:neighbor},
and 
\begin{equation}
h(N,d):=
\begin{cases}
2N^{-1/4}\log(1+N)^{1/2}, & d\le 4,\\[2mm]
2N^{-\frac{1}{4}}+\big(\prod_{i=1}^d(\frac{\sigma_{i, \bm{x}}}{\sigma_{1, \bm{x}}}) N\big)^{-\frac{1}{d}}+\big(\prod_{i=1}^d(\frac{\hat\sigma_{i, \bm{x}}}{\hat\sigma_{1, \bm{x}}}) N\big)^{-\frac{1}{d}}, & d>4.
\end{cases}
\label{hdefine}
\end{equation}
In Eq.~\eqref{hdefine},
\begin{equation}
    \sigma_{i, \bm{x}}\coloneqq \big(\int_{\mathbb{R}^d}y_{i}\mu_{\bm{x}}(\d\bm{y})\big)^{\frac{1}{6}}, \,\, \hat{\sigma}_{i, \bm{x}}\coloneqq \big(\int_{\mathbb{R}^d}y_{i}\hat{\mu}_{\bm{x}}(\d\bm{y})\big)^{\frac{1}{6}}.
\end{equation}
\end{theorem}

The proof of Theorem~\ref{thm:sinkhorn_neighborhood_generalization} is mainly based on the generalization error bound of the local squared $W_2$ distance (\cite[Theorem 2.3]{xia2025efficient}) and is
given in Appendix~\ref{app:sinkhorn_neighborhood_generalization}.
Theorem~\ref{thm:sinkhorn_neighborhood_generalization} provides a generalization error bound on estimating how the empirical Sinkhorn divergence loss function $\bar{S}_{\varepsilon, \delta}^{\text{e}}$, when utilizing the neighborhood technique, approximates the averaged squared $W_2$ distance between the ground truth random field $\bm{y}_{\bm{x}}$ and the approximate model $\hat{\bm{y}}_{\bm{x}}$.

On the other hand, if we use the generalization error bound for estimating the Sinkhorn divergence using the empirical distributions, we can obtain another generalization error bound for our local Sinkhorn divergence loss function.
\begin{theorem}[Generalization bound based on the generalization error of the Sinkhorn divergence]
\rm
\label{thm:neighborhood_sinkhorn_error}
For the averaged local Sinkhorn loss defined in Eq.~\eqref{eq:localsinkhorn} and the corresponding population averaged squared $W_2$ distance defined in Eq.~\eqref{eq:globalsinkhorn}, assume that the hypotheses of \cite[Theorem~3]{genevay2019sample} hold, and $\varepsilon<\frac{2M_0^2}{\sqrt{d}}$.
Then the following generalization error bound holds:
\begin{equation}
\begin{aligned}
\mathbb E\!\left[
\left|
\overline{\mathcal S}_{\varepsilon,\delta}^{\e}(\bm{y}_{\bm{x}},\hat{\bm{y}}_{\bm{x}})
-
\overline{\mathcal W}_{2}(y_x,\hat y_x)
\right|
\right]
\le
2\varepsilon\log\!\Big(
\frac{2e^2M_0^2}{\sqrt d\,\varepsilon}
\Big)+
\mathcal{O}\bigg(
\frac{e^{\frac{5M_0^2}{\varepsilon}}}{\sqrt{N(\bm{x},\delta})}
\Big(1+\varepsilon^{-\lfloor d/2\rfloor}\Big)\bigg)+16M_0^2\delta,
\end{aligned}
\label{eq:neighborhood_sinkhorn_bound}
\end{equation}
where
$N(\bm{x},\delta)$ denotes the effective number of samples in each neighborhood (defined in Eq.~\eqref{N_neighborhood}).

\end{theorem}

The proof of Theorem~\ref{thm:neighborhood_sinkhorn_error} is mainly based on the generalization error bound of using the empirical measures to evaluate the Sinkhorn divergence (\cite[Theorem 3]{genevay2019sample}) and is given in Appendix~\ref{app:sinkhorn_neighborhood_generalization}.


 To exemplify how the size of the neighborhood $\delta$ and the number of samples in each neighborhood $N(\bm{x}, \delta)$ influences the generalization error bound, consider the case when the empirical probability measure of the input variable is close to the uniform distribution and the neighborhood size satisfies $N(\bm{x},\delta)\asymp CN\delta^n$ ($N$ is the total number of training samples) for the intrinsic dimension $n$ of the input variable $\bm{x}$, then using Theorem~\ref{thm:sinkhorn_neighborhood_generalization}, we have:
\begin{equation}
\begin{aligned}
\mathbb E\!\left[
\left|
\overline{\mathcal S}_{\varepsilon,\delta}^{\e}(\bm{y}_{\bm{x}},\hat{\bm{y}}_{\bm{x}})
-
\overline{\mathcal W}_{2}(\bm{y}_{\bm{x}},\hat{\bm{y}}_{\bm{x}})
\right|
\right]
\lesssim
2\varepsilon\log\!\left(
\frac{2e^2M_0^2}{\sqrt d\,\varepsilon}
\right)
+
\frac{4M_0}{\sqrt{N}}
+
8CM_0\,h(CN\delta^n,d)
+16M_0^2\,\delta.
\end{aligned}
\label{eq:neighborhood_sinkhorn_bound_nd0}
\end{equation}

On the other hand, using Theorem~\ref{thm:neighborhood_sinkhorn_error} yields:
\begin{equation}
\begin{aligned}
\mathbb E\!\left[
\left|
\overline{\mathcal S}_{\varepsilon,\delta}^{\e}(\bm{y}_{\bm{x}},\hat{\bm{y}}_{\bm{x}})
-
\overline{\mathcal W}_{2}(\bm{y}_{\bm{x}},\hat{\bm{y}}_{\bm{x}})
\right|
\right]
\lesssim
2\varepsilon\log\!\left(
\frac{2e^2M_0^2}{\sqrt d\,\varepsilon}
\right)+
\mathcal{O}\bigg(
\frac{e^{\frac{5M_0^2}{\varepsilon}}}{\sqrt{CN\delta^n}}
\Bigl(1+\varepsilon^{-\lfloor d/2\rfloor}\Bigr)\bigg) +16M_0^2\delta.
\end{aligned}
\label{eq:neighborhood_sinkhorn_bound_nd}
\end{equation}

The two bounds Eqs.~\eqref{eq:neighborhood_sinkhorn_bound_nd0} and \eqref{eq:neighborhood_sinkhorn_bound_nd} suggest an important practical implication. 
When the data are highly heterogeneous, the conditional distributions may vary sharply across the domain. When the approximate random field model can capture such heterogeneity such that $\prod_{i=1}^d(\frac{\sigma_{i,\bm{x}}}{\sigma_{1, \bm{x}}})^{\frac{1}{d}}+\prod_{i=1}^d(\frac{\hat\sigma_{i,\bm{x}}}{\hat\sigma_{1,\bm{x}}})^{\frac{1}{d}}$ is small, the approximation error induced by overly large entropic regularization when $\varepsilon$ is large can dominate the generalization error. In this regime, it is preferable to choose a relatively small regularization parameter $\varepsilon$, so that the local Sinkhorn loss remains close to the local squared $W_2$ loss. Then, compared with the local squared $W_2$ loss, the additional generalization error induced is essentially only the Sinkhorn regularization bias, while the neighborhood-based approximation and sampling errors remain unchanged.

On the other hand, when the noise in the target variable $\bm{y}_{\bm{x}}$ is homogeneous, the conditional distributions vary more smoothly and a moderately large regularization parameter, e.g. $\varepsilon=O(1)$ at the scale of the cost function, can be advantageous, so that the convergence rate as the number of training samples $N\rightarrow\infty$ is $\mathcal{O}\big((N\delta^{n})^{-\frac{1}{2}}\big)$. In this case, entropic smoothing improves numerical stability and can reduce the effective sample complexity penalty associated with the Sinkhorn approximation. As a result, by tuning $\varepsilon$, the local Sinkhorn loss provides a flexible trade-off between statistical accuracy and computational robustness, and may partially alleviate the curse of dimensionality compared with the exact local squared $W_2$ loss.  Furthermore, as $\varepsilon$ decreases, the proposed local Sinkhorn loss approaches the local squared Wasserstein distance, leading to higher geometric fidelity but slower statistical convergence. Conversely, increasing $\varepsilon$ improves the statistical convergence and computational efficiency of the empirical Sinkhorn divergence while introducing additional regularization bias. Therefore, selecting a moderate regularization parameter also provides a favorable compromise between approximation accuracy and computational efficiency.

Finally, the neighborhood radius $\delta$ must be chosen to balance two competing sources of error. As shown in Eqs.~\eqref{eq:neighborhood_sinkhorn_bound_nd0} and \eqref{eq:neighborhood_sinkhorn_bound_nd}, increasing $\delta$ enlarges the number of samples within each neighborhood and therefore reduces the empirical estimation error, namely
$8CM_0\,h(CN\delta^n,d)$ and
$\mathcal{O}\!\left(
\frac{\exp\!\left(5M_0^2/\varepsilon\right)}
{\sqrt{CN\delta^n}}
\left(1+\varepsilon^{-\lfloor d/2\rfloor}\right)
\right)$.
However, a larger neighborhood inevitably increases the systematic approximation error, quantified by the term $16M_0^2\delta$. Consequently, an appropriate neighborhood radius should strike a balance between statistical accuracy and locality: it should not be too small to provide adequate samples for reliable estimation of the local empirical distribution, while remaining not too large to preserve the local structure of the underlying conditional distribution and avoid excessive approximation bias.


\section{Numerical experiments}
\label{section3}
In this section, we carry out numerical experiments to test our proposed local Sinkhorn approach for training SNNs. Test settings and hyperparameters are listed in Table~\ref{tab:hyperparameters}.  Pseudocode for minimizing our proposed local Sinkhorn to train the SNN in Fig.~\ref{fig:snn} is given in Algorithm~\ref{alg:local_sinkhorn_style}.

\begin{algorithm}[t]
\caption{The pseudocode of the local Sinkhorn approach to train an SNN}
\label{alg:local_sinkhorn_style}
\begin{algorithmic}[1]
\Require Given $N$ observed data $\{(\bm{x}_i,\bm{y}_{\bm{x}_i})\}_{i=1}^{N}$, the neighborhood radius $\delta$, the mini-batch size $n_b$, the number of local samples $n$, the minimum number of local samples required for each center $n_{\min}$, the maximal number of local samples recorded for each center $n_{\max}$,
the number of epochs $E$, and the Sinkhorn regularization parameter $\varepsilon$
\State Initialize the SNN in Fig.~\ref{fig:snn}
\State For each $x_i$, find its neighborhood $ B_i\coloneqq\{\bm{x}_j:\|\bm{x}_j-\bm{x}_i\|\le \delta\}$. Do not use $\bm{x}_i$ as an eligible training neighborhood center if its neighborhood contains too few samples ($|B_i|< n_{\min}$). If there are too many neighbors such that $|B_i|>n_{\max}$, randomly keep $n_{\max}$ neighbors to form the updated $B_i$
\State Input $\{\bm{x}_i\}_{i=1}^{N}$ into the SNN model to obtain predictions $\{\hat{\bm{y}}_i\}_{i=1}^{N}$
\For{$\mathrm{epoch}=1,\ldots,E$}
    \State Randomly select $n_b$ neighborhood centers
    \For{each selected center $\bm{x}_i$}
        \State Randomly choose $\min(n, |B_i|)$ samples from $ B_i$ to construct a local mini-batch ($\mathcal{B}(\bm{x}_i,\delta)$ defined in Eq.~\eqref{tilde_B_def})
        \State Input the sampled local inputs into the SNN to obtain predictions
        \State Compute the local Sinkhorn loss
        \[
        \mathcal L_i = S_\varepsilon\!\left(\hat \mu_{\bm{x}_i,\delta}^{\text{e}},\mu_{\bm{x}_i,\delta}^{\text{e}}\right)
        \]
    \EndFor
    \State Calculate the total loss
    \[
    \mathcal L=\frac{1}{n_b}\sum_{i=1}^{n_b}\mathcal L_i
    \]
    \State Perform gradient descent to minimize $\mathcal L$ and update the parameters in the SNN
    \State Resample the stochastic weights in the SNN using the updated parameter distributions
\EndFor
\State \Return The trained SNN
\end{algorithmic}
\end{algorithm}

In Examples \ref{example1} and 2, the reconstruction accuracy on the testing set is evaluated by comparing the empirical conditional means and variances of the reconstructed and reference distributions. Let $\{(\bm{x}_i^{\text{test}}, \bm{y}_i^{\text{test}})\}$ denote samples in the testing set, and $B^{\text{test}}(\bm{x}_i^{\text{test}},0)$ denote the set of observations with the identical input as $\bm{x}_i$:
\begin{equation}
\{\bm{x}_j^{\text{test}}: \bm{x}_j^{\text{test}} = \bm{x}_i^{\text{test}}\}.
\end{equation}
The ground-truth and predicted means are:
\begin{equation}
\bm{m}_i
=
\frac{1}{R}
\sum_{\bm{x}_j^{\text{test}} \in B(\bm{x}_i^{\text{test}},0)}
\bm{y}_j^{\text{test}},
\qquad
\hat{\bm{m}}_i
=
\frac{1}{R}
\sum_{\bm{x}_j^{\text{test}} \in B(\bm{x}_i^{\text{test}},0)}
\hat{\bm{y}}_j^{\text{test}}
\label{mean_def}
\end{equation}
be the empirical conditional mean vectors at the $i^{\text{th}}$ testing center, respectively. 
Similarly, let
\begin{equation}
\bm{v}_i
=
\frac{1}{R}
\sum_{\bm{x}_j^{\text{test}} \in B(\bm{x}_i^{\text{test}},0)}
\left(
\bm{y}_j^{\text{test}}-\bm{m}_i
\right)^{\circ2},
\qquad
\hat{\bm{v}}_i
=
\frac{1}{R}
\sum_{\bm{x}_j^{\text{test}} \in B(\bm{x}_i^{\text{test}},0)}
\left(
\hat{\bm{y}}_j^{\text{test}}-\hat{\bm{m}}_i
\right)^{\circ2},
\label{var_def}
\end{equation}
denote the empirical variance vectors at the $i^{\text{th}}$ testing center, where $(\cdot)^{\circ2}$ denotes the element-wise square. In Eqs.~\eqref{mean_def} and \eqref{var_def}, $R\coloneqq |B^{\text{test}}(\bm{x}_i^{\text{test}},0)|$ is the number of testing samples at each testing center.
The average relative mean and variance reconstruction error is defined as follows:
\begin{equation}
\mathcal E_{\mathrm{mean}}
=
\frac1{N_c}
\sum_{i=1}^{N_c}
\frac{
\|
\hat{\bm{m}}_i-\bm{m}_i
\|_2
}
{
\|
\bm{m}_i
\|_2+\epsilon_0
},\,\,
\mathcal E_{\mathrm{var}}
=
\frac1{N_c}
\sum_{i=1}^{N_c}
\frac{
\|
\hat{\bm{v}}_i-\bm{v}_i
\|_2
}
{
\|
\bm{v}_i
\|_2+\epsilon_0
},
\end{equation}
where $N_c$ is the number of testing centers, and $\epsilon_0$ is set as $10^{-8}$ to avoid division by zero.

First, we consider learning a unidimensional random field model to compare with different loss functions.

\begin{example}
\rm
\label{example1}


We consider a one-dimensional conditional distribution reconstruction problem. The objective is to learn the conditional distribution
\begin{equation}
p(y\,|\,x), \,\,x\in[0,1].
\end{equation}
Unlike conventional regression problems where only the conditional mean is estimated, our objective is to reconstruct the entire conditional distribution from scattered observations.
The conditional response is generated from a bimodal Gaussian mixture model,
\begin{equation}
y=m(x)+s\,d(x)+\varepsilon_0,
\label{eq:example1_model}
\end{equation}
where the random variable $s\in\{-1,+1\}$
is sampled with equal probability,
\[
\mathbb P(s=-1)=\mathbb P(s=1)=\frac12,
\]
and 
$\varepsilon_0\sim\mathcal N(0,\sigma^2)$ with $\sigma=0.04$. In Eq.~\eqref{eq:example1_model}, the conditional mean function is chosen as
\begin{equation}
m(x)
=
0.50
+0.20x
+\exp\!\left[-5(x-0.60)^2\right]
+0.40\sin\!\left(\frac{x}{2}\right),
\label{eq:example1_mean}
\end{equation}
which produces a smooth nonlinear trend over the computational domain. The separation between the two Gaussian modes is controlled by
\begin{equation}
d(x)
=
0.38
+
0.10
\exp\!\left(
-\frac{(x-0.70)^2}
{2\cdot(0.14)^2}
\right),
\label{eq:example1_sep}
\end{equation}
which allows the distance between the two modes to vary smoothly with the input variable.
Consequently, the conditional distribution is given by
\begin{equation}
p(y|x)
=
\frac12
\mathcal N
\left(
m(x)-d(x),
\sigma^2
\right)
+
\frac12
\mathcal N
\left(
m(x)+d(x),
\sigma^2
\right).
\label{eq:example1_pdf}
\end{equation}

For the training set, $N_{\rm train}=2000$ samples are independently generated by first sampling
$x_i\sim \mathcal{U}(0,1)$,
followed by drawing
$y_i\sim p(y|x_i)$.
Therefore, each training sample is independent. To generate the test set, $100$ uniformly distributed input locations are selected as neighborhood centers. At each testing location, $20$ independent realizations are generated from the conditional distribution (Eq.~\ref{eq:example1_pdf}), providing empirical estimates of the local conditional distributions used for quantitative evaluation.

\begin{figure}[h]
\centering
\includegraphics[width=\textwidth]{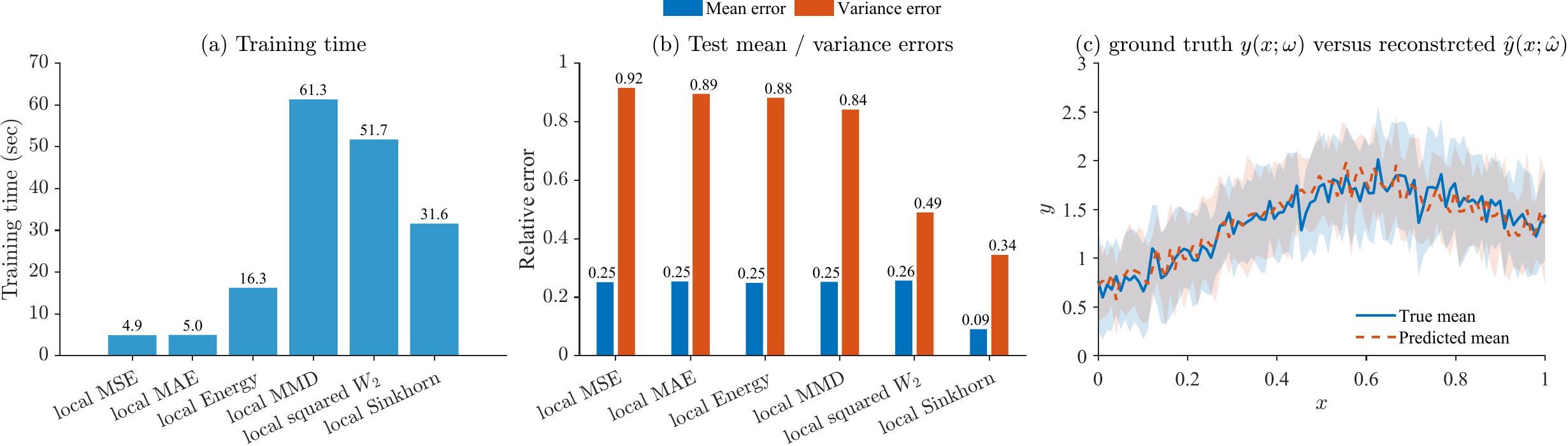}
\caption{
Comparison of different local distribution matching losses for Example~\ref{example1}
(a) Total training time.
(b) Average relative errors of the conditional mean and conditional variance on the testing set.
(c) Comparison between the true and predicted conditional distributions obtained using the proposed Local Sinkhorn loss. The solid curves denote the conditional means, while the shaded regions represent one standard deviation. All loss functions are minimized to train the same SNN architecture and identical training hyperparameters.
}
\label{fig:example1}
\end{figure}

To investigate the influence of different discrepancy measures on conditional distribution learning, six different local loss functions are considered, namely local MSE, local MAE, local Energy Distance, local MMD, local squared $W_2$, and the proposed local Sinkhorn divergence. The mathematical definitions of these losses are summarized in Appendix~\ref{app:loss}.
Fig.~\ref{fig:example1} summarizes the overall performance of the six loss functions. Fig.~\ref{fig:example1} (a) compares the total training time. Owing to the entropic regularization, the proposed Local Sinkhorn loss requires significantly less computational effort than the previous local squared $W_2$ loss in \cite{xia2026local}.
Fig.~\ref{fig:example1} (b) reports the average relative errors of the conditional mean and conditional variance on the testing set. Distribution-based losses consistently outperform pointwise regression losses (Local MSE and Local MAE), indicating that matching local empirical distributions is substantially more effective than matching individual samples. Among all competing losses, the proposed local Sinkhorn loss achieves the smallest overall errors for both the conditional mean and conditional variance.
Fig.~\ref{fig:example1} (c) compares the reconstructed conditional distributions obtained by the proposed Local Sinkhorn model with the reference solution. The predicted mean accurately follows the true nonlinear trend while the predicted standard-deviation band agrees closely with the reference distribution over the entire computational domain. These results demonstrate that the proposed local Sinkhorn formulation is capable of accurately reconstructing both the location and spread of the conditional distribution.
\end{example}

Next, we carry out an experiment in which the target random variable $\bm{y}_{\bm{x}}$ is multidimensional and compare our local Sinkhorn approach with other machine-learning uncertainty quantification benchmarks.

\begin{example}
    \rm
    \label{example2}

We consider a stochastic Darcy flow problem on the spatial domain $\bm{x}\in\mathcal{D}=[0,1]^2$. For each realization of the uncertain model parameters $\omega\in\Omega$, we solve:
\begin{equation}
-\nabla \cdot \left(a(\bm{x},\omega)\nabla u(\bm{x},\omega)\right)
=
f(\bm{x}),
\qquad \bm{x}\in D,
\end{equation}
subject to homogeneous Dirichlet boundary conditions
\begin{equation}
u(\bm{x},\omega)=0,
\qquad \bm{x}\in\partial D.
\end{equation}

The permeability field is modeled as:
\begin{equation}
a(\bm{x},\omega)=\exp\!\left(g(\bm{x},\omega)\right),
\label{eq:permeability}
\end{equation}
where $g(\bm{x},\omega)$ is a zero-mean Gaussian random field generated by FFT-based spectral synthesis. Specifically, the random field is constructed in the Fourier domain as:
\begin{equation}
\widehat g(k,\omega)
=
\sqrt{c(k)}\,\widehat\xi(k,\omega),
\label{eq:fft_grf}
\end{equation}
where $\widehat\xi(k,\omega)$ are independent complex Gaussian random variables satisfying $
\widehat\xi(k,\omega)\sim\mathcal N(0,1)$,
and the covariance spectrum is prescribed by
\begin{equation}
c(k)
=
\left(
1+
(2\pi\ell)^2|k|^2
\right)^{-\alpha/2},
\label{eq:spectrum}
\end{equation}
with correlation length $\ell$ and smoothness parameter $\alpha$. The Gaussian random field is then obtained by applying the inverse Fourier transform,
\begin{equation}
g(\bm{x},\omega)
=
\mathcal F^{-1}
\!\left(
\widehat g(k,\omega)
\right),
\label{eq:ifft}
\end{equation}
followed by normalization to zero mean and unit variance.


To generate the training and testing set, we establish a uniform $64\times 64$ grid on $\mathcal{D}$ denoted as $\mathcal{D}_{64}=\{\bm{x}_{i, j}=(\frac{i}{63}, \frac{j}{63})\}_{i,j=0}^{64}$. Then, we randomly sample $\bm{x}_{i, j}$ from $\mathcal{D}_{64}$ satisfying $6\leq i, j\leq 57$. For the chosen $\bm{x}_{i, j}$, a square window of size
$7\times 7$ is first constructed on the underlying computational grid.
A total of $16$
grid points $\bm{x}_{i', j'}$ are then randomly sampled without replacement from this window. This sampling strategy is adopted to ensure that most training samples have enough neighbors around them. 
For each selected grid point $\bm{x}_{i', j'}$, a
$5\times 5$
solution patch on the grid is extracted with the selected grid point being the center. 
The corresponding solution patch is defined as:
\begin{equation}
\bm{y}(\bm{x}_{i', j'},\omega_{i', j'})
=
\Big(
u(\bm{x}_{i'-2, j'-2},\omega_{i'-2, j'-2}),
u(\bm{x}_{i'-2, j'-1},\omega_{i'-2, j'-1}),
\ldots,
u(\bm{x}_{i'+2, j'+2},\omega_{i'+2, j'+2})
\Big)^\top
\in\mathbb R^{25}.
\end{equation}
An illustration of the training and testing set sample generation strategy is given in Fig.~\ref{fig:sensitivity} (a).
The vector $\bm{y}(\bm{x},\omega)$ serves as the target random field whose conditional distribution is to be reconstructed when $\bm{x}$ is given. To generate the testing set, 20 testing centers are generated. For each testing center $\bm{x}_{\ell}^{\mathrm{test}}$, a large number of independent permeability realizations $\omega_1,\ldots,\omega_M$ are generated.
The resulting collection
\begin{equation}
\left\{
\bm{y}(\bm{x}_{\ell}^{\mathrm{test}},\omega_{\ell})
\right\}_{j=1}^{M}
\end{equation}
provides an empirical approximation of the conditional distribution
$p\!\left(
\bm{y}
\mid
\bm{x}_{\ell}^{\mathrm{test}}
\right)$. The learned model is then evaluated by comparing the predicted conditional distribution with this empirical reference distribution at the test locations. This benchmark naturally produces multidimensional output variables with strong spatial correlations, making it a suitable experiment to test if our local Sinkhorn approach could successfully reconstruct a multidimensional random field in which the noise is strongly heterogeneous or lies in a low-dimensional manifold.

To evaluate the effectiveness of the proposed local Sinkhorn divergence, we compare it with several representative approaches for conditional distribution reconstruction. 
The compared loss functions include local Maximum Mean Discrepancy (MMD) and the local squared $W_2$ loss proposed in our previous work \cite{xia2026local}, whose definitions are detailed in Appendix~\ref{app:loss}. To further compare with representative probabilistic learning approaches, we also consider Heteroscedastic Gaussian Regression, Mixture Density Networks (MDN) \cite{bishop1994mixture}, Conditional Variational Autoencoders (CVAE) \cite{sohn2015learning}, and Conditional Normalizing Flows (CNF) \cite{rezende2015variational,dinh2017density,papamakarios2021normalizing}. These methods represent commonly used likelihood-based and latent-variable models for conditional distribution estimation. All benchmark methods are trained and evaluated using the same datasets and testing protocols to ensure a fair comparison.




\begin{table*}[ht]
\footnotesize
\centering
\caption{Comparison of different conditional distribution learning methods and loss function used to train the SNN on the Darcy flow benchmark.}
\label{tab:benchmark}
\renewcommand{\arraystretch}{1.2}
\begin{tabular}{lcccccc}
\hline
Method &
Mean Error &
Variance Error &
Training Time (s) &
Memory (MB) \\
\hline

Heteroscedastic Gaussian Regression
& 0.1189
& 0.5104
& 23.48
& 687.38 \\

Mixture Density Network
& 0.0970
& 5.3101
& 52.06
& 686.72 \\

Conditional VAE
& 0.2931
& 5.8687
& 31.10
& 695.09 \\

Conditional Normalizing Flow
& 0.3256
& 3.5312
& 140.36
& 689.73 \\

\hline



SNN + Local $W_2$
& 0.0519
& 0.3213
& 500.84
& 905.14 \\

SNN + Sinkhorn
& 0.0473
& 0.2318
& 308.34
& 791.08 \\

SNN + MMD
& 0.0779
& 0.7878
& 649.60
& 733.90 \\

\hline
\end{tabular}
\end{table*}

Table~\ref{tab:benchmark} compares the proposed neighborhood-based distribution learning methods with several representative conditional generative baselines.
Among the likelihood-based models, the heteroscedastic Gaussian regression is computationally the most efficient, requiring only $23.48$ seconds for training, but its Gaussian assumption limits the reconstruction accuracy, especially for the variance. MDN improves the mean prediction accuracy but still suffers from poor variance reconstruction. CVAE and CNF provide more expressive conditional generative models, although their reconstruction errors remain significantly larger than those of the neighborhood-based approaches.

The neighborhood-based optimal transport methods consistently outperform the global conditional generative models. In particular, training the SNN with the local Sinkhorn divergence achieves the smallest errors in the predicted mean and variance. Furthermore, compared to the previous local squared $W_2$ approach, its computational efficiency is improved, and the runtime is shortened. The proposed neighborhood-based Sinkhorn approach to train SNN provides the best overall trade-off between reconstruction accuracy and computational efficiency. These results suggest that once local conditional distributions are exploited with the neighborhood technique, the simple SNN can be efficiently trained using our local Sinkhorn loss, which is sufficient to outperform considerably more sophisticated machine-learning conditional generative models. Overall, the experimental results demonstrate that exploiting local neighborhoods when utilizing the OT-based loss functions is considerably more important than increasing the complexity of the conditional generator.

To investigate the robustness of the proposed local Sinkhorn framework with respect to different algorithmic parameters, we further perform sensitivity studies on the noise level of the training data, the neighborhood radius, and the Sinkhorn regularization parameter. The corresponding results are presented in Fig.~\ref{fig:sensitivity}. For the noise sensitivity study, we vary the intensity of the underlying Gaussian random field. The permeability field is constructed from
\begin{equation}
g_\sigma(\bm{x},\omega)
=
\sigma\, g(\bm{x},\omega),
\label{eq:noiselevel}
\end{equation}
where $\sigma>0$ is referred to as the noise level and $g$ denotes the normalized Gaussian random field generated by the FFT-based spectral synthesis in Eq.~\eqref{eq:ifft}. Consequently, the permeability field is given by
\[
a(\bm{x},\omega)
=
\exp\!\left(g_\sigma(\bm{x},\omega)\right).
\]
Increasing $\sigma$ enlarges the variance of the permeability field and therefore increases the intrinsic uncertainty of the corresponding Darcy solutions.

\begin{figure}[h]
\centering
\includegraphics[width=\textwidth]{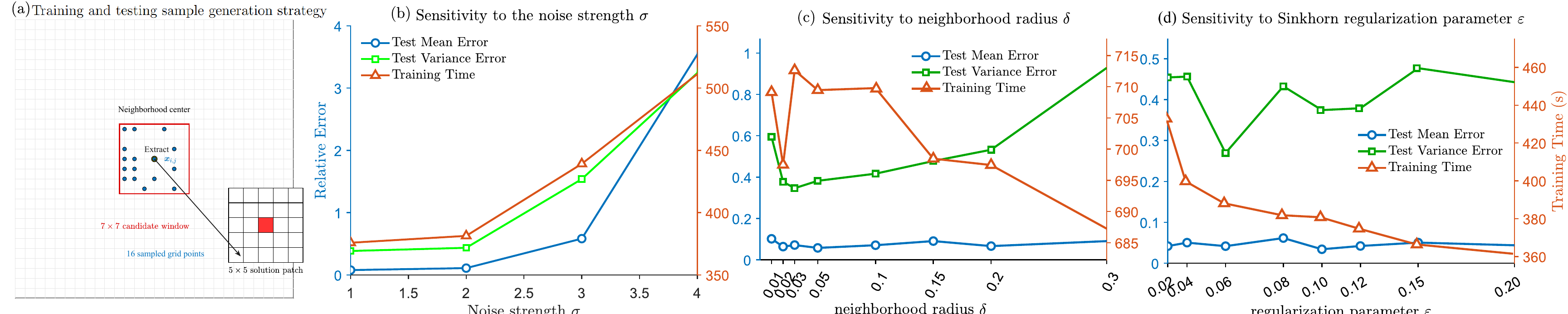}
\caption{
(a) An illustration of the training and testing sample selection strategy. 
(b) Influence of the random-field intensity parameter $\sigma$, where the Gaussian random field is generated as
$g_\sigma(\bm{x},\omega)=\sigma g_0(\bm{x},\omega)$.
(c) Influence of the neighborhood radius $\delta$ used to construct local empirical conditional distributions.
(d) Influence of the Sinkhorn regularization parameter $\varepsilon$. For all panels, the left axis reports the average relative reconstruction errors of the conditional mean and variance, while the right axis reports the corresponding training time.
}
\label{fig:sensitivity}
\end{figure}

Fig.~\ref{fig:sensitivity} (b) illustrates the influence of the observation noise level. As expected, increasing the noise level deteriorates the reconstruction accuracy of both the conditional mean and variance, while the training time increases only moderately. This indicates that the proposed method remains computationally stable even under relatively noisy observations.
Fig.~\ref{fig:sensitivity} (c) studies the effect of the neighborhood radius $\delta$. A sufficiently small neighborhood does not contain enough local samples to accurately approximate the conditional distribution, whereas an excessively large neighborhood violates the locality assumption and mixes samples associated with different conditional distributions. Consequently, an intermediate neighborhood radius provides the best trade-off between approximation accuracy and computational cost.
Fig.~\ref{fig:sensitivity} (d) shows the sensitivity of the proposed method to the Sinkhorn regularization parameter $\varepsilon$. When $\varepsilon$ is too small, the Sinkhorn divergence approaches the exact Wasserstein distance. In this regime, the reconstructed variance error is large because the empirical squared $W_2$ distance converges slowly to the ground truth squared $W_2$ distance in multidimensional settings. Furthermore, choosing a too small $\varepsilon$ leads to longer runtime. 
Conversely, a large regularization parameter introduces excessive entropic smoothing and leads to larger reconstruction errors. These observations agree well with the theoretical error analysis in Section~\ref{section2}, where the regularization parameter controls the trade-off between approximation bias and computational efficiency, and a moderate regularization parameter $\varepsilon$ is preferable to reconstruct multidimensional random field models.
Overall, the proposed local Sinkhorn framework exhibits good robustness over a reasonably wide range of neighborhood radii and regularization parameters.
\end{example}

Finally, we consider reconstructing a stochastic dynamical system using our proposed local Sinkhorn approach.

\begin{example}
\rm
\label{example3}

To further demonstrate the applicability of the proposed method to multidimensional stochastic dynamical systems, we consider a network of coupled nonlinear stochastic FHN oscillators in computational neuroscience \cite{Tuckwell1998, Acebron2004}.

We consider a network consisting of $N_n=5$ coupled neurons, resulting in a $10$-dimensional stochastic differential equation. Let
\[
\bm{x}(t;\bm{x}_0;\omega)
=
\big(
v_1(t),\ldots,v_5(t),
w_1(t),\ldots,w_5(t)
\big)^\top
\in\mathbb{R}^{10},
\]
where $v_i$ denotes the membrane potential and $w_i$ is the corresponding recovery variable.
The governing stochastic system is
\begin{equation}
\begin{aligned}
\d v_i(t)
&=
\left(
v_i
-\frac{v_i^3}{3}
-
w_i
+
\sum_{j=1}^{5}
C_{ij}(\omega)v_j
\right)\d t
+
\sigma_i(\omega)\,\d B_i(t),
\\
\d w_i(t)
&=
\varepsilon_i(\omega)
\left(
v_i+c-d w_i
\right)\d t,
\qquad
i=1,\ldots,5, \,\, (v_1(0),\ldots, v_5(0), w_1(0), \ldots, w_5(0)) = \bm{x}_0,
\end{aligned}
\label{eq:fhn}
\end{equation}
where $B_i(t)$ are independent standard Brownian motions, and $c=0.7$ and $d=0.8$ are fixed deterministic parameters.

Unlike the classical FHN model, the coupling strengths
$C_{ij}(\omega)$, the diffusion coefficients 
$\sigma_i(\omega)$,
and the recovery time scales
$\varepsilon_i(\omega)$
are all regarded as random variables whose values may vary across different realizations, and $\omega$ denotes the uncertainty in those model parameters. More precisely,
$
\left(
C_{ij}(\omega),
\sigma_i(\omega),
\varepsilon_i(\omega)
\right)$
is independently sampled for every realization, yielding a stochastic drift field
$a(x,\omega)$
and a stochastic diffusion field
$b(x,\omega)$. Consequently, the dynamics can be written compactly as
\begin{equation}
\d \bm{x}(t;\bm{x}_0;\omega)
=
a(\bm{x}(t;\bm{x}_0;\omega),\omega)\,\d t
+
b(\bm{x}(t;\bm{x}_0;\omega),\omega)\,\d \bm{B}_t, \,\, \bm{x}(0;\bm{x}_0;\omega) = \bm{x}_0, \,\, t\in[0, T],
\label{eq:compact_fhn}
\end{equation}
where both the drift and diffusion are random functions induced by the uncertain parameters,
$\bm{B}_t$ is a five-dimensional standard Brownian motion, and
$\bm{x}_0$ denotes the initial condition, with
$\bm{x}(t;\bm{x}_0;\omega)$ explicitly indicating the dependence of the solution on $\bm{x}_0$.


The reference trajectories and predicted trajectories are generated using the Euler--Maruyama solver implemented in the \texttt{torchsde} package. For the training set, each trajectory is associated with an independently sampled initial condition and an independent realization of the random parameter vector $\omega$. The resulting dataset consists of 
$\Big\{\big(\bm{x}_0^{(i)},\bm{x}^{(i)}(t_1;\bm{x}_0^{(i)};\omega^{(i)}),\ldots,\bm{x}^{(i)}(t_{N_t};\bm{x}_0^{(i)};\omega^{(i)})\big)\Big\}_{i=1}^{N_{\mathrm{train}}}$, where $\bm{x}^{(i)}(t_k)\in\mathbb{R}^{10}$ is the snapshot of the trajectory at time $t_k = \frac{kT}{N_t}$.

For the testing set, we randomly select $20$ initial conditions. For each initial condition, $20$ independent realizations of the random parameters are generated, producing empirical conditional distributions of the solution trajectories. Therefore, the testing data naturally approximates the conditional distribution
$P
\left(
\bm{x}^{\text{test}}(t;\bm{x}_0;\omega)
\mid
\bm{x}^{\text{test}}(0;\bm{x}_0;\omega)=\bm{x}_0
\right)$,
which serves as the reference distribution for evaluating the reconstructed stochastic dynamics. For both the training set and the testing set, we set the time horizon $T=1$ and the time steps $N_t=10$ in Eq.~\eqref{eq:compact_fhn}.

To reconstruct the unknown stochastic dynamics, we construct the approximate model:
\begin{equation}
\d \hat{\bm{x}}(t; \bm{x}_0; \hat{\omega})
=
\hat{a}(\hat{\bm{x}}(t; \bm{x}_0; \hat{\omega}),\hat{\omega}_a)\,\d t
+
\hat{b}(\hat{\bm{x}}(t; \bm{x}_0; \hat{\omega}),\hat{\omega}_b)\,\d B_t, \,\, \hat{\bm{x}}(0;\bm{x}_0; \hat{\omega}) = \bm{x}_0,
\label{eq:compact_fhn_approximate}
\end{equation}
where $\hat a(\bm{x}(t; \bm{x}_0; \hat{\omega}),t;\hat{\omega}_a)$ and $\hat b(\bm{x}(t; \bm{x}_0; \hat{\omega}),t;\hat{\omega}_b)$ are two separate SNNs and $\hat{\omega}=(\hat{\omega}_a,\hat{\omega}_b)$ refers to the union of uncertain weights in $\hat{a}$ and $\hat{b}$.
$\hat{a}$ and $\hat{b}$ in Eq.~\eqref{eq:compact_fhn_approximate} take the 10-dimensional state variable $\hat{\bm{x}}(t;\bm{x}_0; \hat{\omega})$ and time $t$ as the input and then output 10-dimensional approximate drift and diffusion functions to approximate the stochastic drift and diffusion coefficients in Eq.~\eqref{eq:compact_fhn}, respectively. Since the dynamics of $w_i$ in Eq.~\eqref{eq:fhn} are noise-free, we enforce the last five components of $\hat{b}$ in Eq.~\eqref{eq:compact_fhn_approximate} to be 0.
During one trajectory simulation, a single realization of the network weights $\hat{\omega}$ is sampled and kept fixed throughout the entire time interval. Consequently, the randomness remains trajectory-wise rather than time-wise, which is consistent with the assumption that the underlying uncertain parameters are realization-dependent. The two SNNs are trained simultaneously by minimizing a temporal-average local Sinkhorn loss:
\begin{equation}
\begin{aligned}
    \frac{1}{T}\int_0^T S_{\varepsilon, \delta}^{\text{e}} (\bm{x}\big(t;\bm{x}_0;\omega), \hat{\bm{x}}(t;\bm{x}_0;\hat\omega)\big)\d t&\approx \frac{1}{N_T}\sum_{i=1}^{N_T}\overline{S}^{\text{e}}_{\varepsilon, \delta}(\bm{x}\big(\tfrac{iT}{N_T};\bm{x}_0;\omega), \hat{\bm{x}}(\tfrac{iT}{N_T};\bm{x}_0;\hat\omega)\big)\\&=\frac{1}{N_T}\sum_{i=1}^{N_T}\int_{\mathcal{D}} S_{\varepsilon}\big(\mu^{\text{e}}_{\bm{x}_0, \delta}(\tfrac{iT}{N_T}), \hat{\mu}^{\text{e}}_{\bm{x}_0, \delta}(\tfrac{iT}{N_T})\big)\nu^{\text{e}}(\d\bm{x}_0),
    \end{aligned}
    \label{temporal_avarage}
\end{equation}
where $\bm{x}_0$ represents the initial condition, and
$\mu^e_{\bm{x}_0,\delta}(t)$ and
$\widehat{\mu}^e_{\bm{x}_0,\delta}(t)$ denote the empirical distributions
of $\bm{x}(t;\bm{x}_0;\omega)$ and
$\widehat{\bm{x}}(t;\bm{x}_0;\widehat{\omega})$, respectively,
constructed from trajectories whose initial conditions fall within the selected neighborhood subset
$\mathcal{B}(\bm{x}_0,\delta)\subseteq B(\bm{x}_0,\delta)$.

 The proposed temporal-average local Sinkhorn loss Eq.~\eqref{temporal_avarage} compares the empirical distributions of the predicted and reference trajectories at every observation time and then averages the discrepancies over the entire temporal domain. Compared with the temporal-average local squared $W_2$ loss introduced in our previous work \cite{xia2026local}, which replaces the Sinkhorn divergence with the squared $W_2$ distance in Eq.~\eqref{temporal_avarage}, the proposed Sinkhorn formulation provides a differentiable approximation with substantially improved computational efficiency while preserving the geometric structure of optimal transport. 

\begin{figure}[h]
\centering
\includegraphics[width=0.9\textwidth]{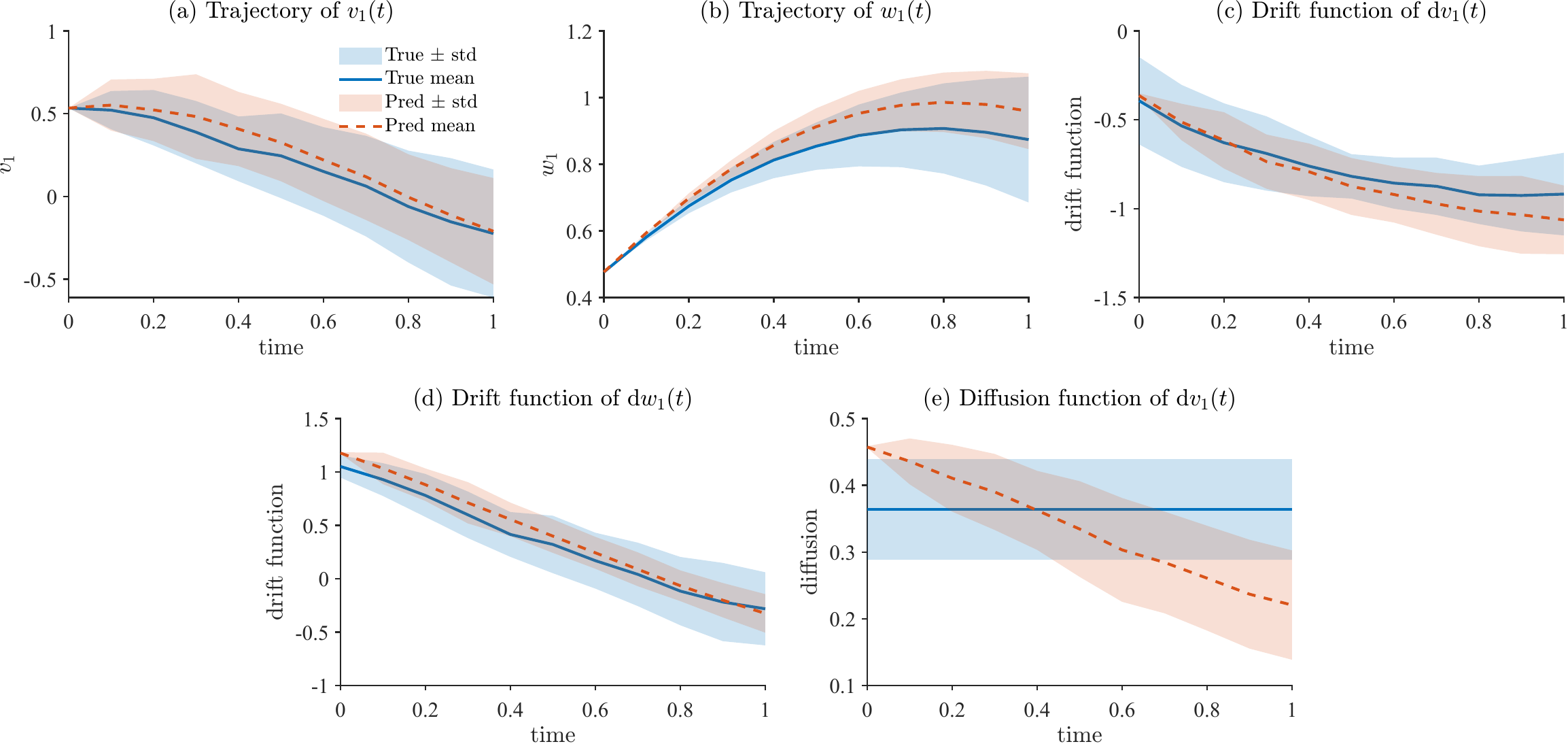}
\caption{
Comparison between the true and predicted stochastic dynamics for a representative testing sample in the stochastic FHN system.
Panels (a)(b) show the mean and standard deviation of trajectories of $v_1(t)$ and $w_1(t)$. (c)(d) show the mean and standard deviation of the drift functions of $\d v_1(t)$ and $\d w_1(t)$ in Eq.~\eqref{eq:fhn}. (e) shows the mean and standard deviation of the diffusion functions of $\d v_1(t)$ in Eq.~\eqref{eq:fhn}.
The shaded regions indicate one standard deviation estimated from multiple stochastic realizations.
}
\label{fig:fhn_prediction}
\end{figure}

To further evaluate the reconstruction quality of the proposed method, Fig.~\ref{fig:fhn_prediction}  compares the predicted stochastic dynamics with the corresponding reference solution for a representative testing sample. In addition to the stochastic trajectories, the learned drift and diffusion functions are also compared with the ground truth. Since the proposed model reconstructs conditional probability distributions rather than individual trajectories, the conditional mean and one-standard-deviation bands estimated from multiple stochastic realizations are reported.
As shown in Fig.~\ref{fig:fhn_prediction} (a)(b), the predicted trajectories accurately reproduce both the temporal evolution and the associated uncertainty of the stochastic FHN system. For both the membrane potential variable $v_1$ and the recovery variable $w_1$, the predicted conditional means closely follow the reference solution, while the predicted uncertainty bands exhibit good agreement with the corresponding ground-truth distributions. Furthermore, the learned drift functions capture the nonlinear deterministic dynamics with high accuracy (Fig.~\ref{fig:fhn_prediction} (c)(d)), and the diffusion functions provide reasonable approximations of the stochastic forcing, while larger errors in the predicted diffusion functions may be due to error accumulation (shown in Fig.~\ref{fig:fhn_prediction} (e)). 
Overall, these results demonstrate that the proposed local Sinkhorn divergence is capable of simultaneously reconstructing the deterministic and stochastic components of the underlying dynamics from scattered conditional observations. 

We also compare minimizing our temporal-average local Sinkhorn divergence Eq.~\eqref{temporal_avarage} with minimizing the previous temporal-average local squared $W_2$ loss in \cite{xia2026local}. The following error metrics are used to quantify the errors of $\hat{\bm{x}}$, $\hat{{a}}$, and $\hat{{b}}$ of the reconstructed SDE \eqref{eq:compact_fhn_approximate} on the testing set, respectively:
    \begin{equation}
\begin{aligned}
        &\text{error in}~\hat{\bm{x}}\coloneqq \frac{\sum_{j=1}^{N_{\text{test}}}\sum_{i=1}^{N_T} W_2^2\big(\mu^{\text{e}}_{\bm{x}_{0, j}^{\text{test}}}(\tfrac{iT}{N_T}), \hat{\mu}^{\text{e}}_{\bm{x}_{0, j}^{\text{test}}}(\tfrac{iT}{N_T})\big)}{\sum_{j=1}^{N_{\text{test}}}\sum_{i=1}^{N_T} \E\Big[\big\|\bm{x}(\tfrac{iT}{N_T};\bm{x}_{0, j}^{\text{test}};\omega)\big\|^2\Big]} ,\,\,\,\,\,\,
    \text{error in}~\hat{a}\coloneqq \frac{\sum_{j=1}^{N_{\text{test}}}\sum_{i=1}^{N_T} W_2^2\big(\eta^{\text{e}}_{a, \bm{x}_{0, j}^{\text{test}}}(\tfrac{iT}{N_T}), \hat{\eta}^{\text{e}}_{a, \bm{x}_{0, j}^{\text{test}}}(\tfrac{iT}{N_T})\big)}{\sum_{j=1}^{N_{\text{test}}}\sum_{i=1}^{N_T} \E\Big[\big\|{a}(\bm{x}(\tfrac{iT}{N_T};\bm{x}_{0, j}^{\text{test}};\omega),\omega)\big\|^2\Big]},\\
&\qquad\qquad     \text{error in}~\hat{b}\coloneqq \frac{\sum_{j=1}^{N_{\text{test}}}\sum_{i=1}^{N_T} W_2^2\big(\eta^{\text{e}}_{b, \bm{x}_{0, j}^{\text{test}}}(\tfrac{iT}{N_T}), \hat{\eta}^{\text{e}}_{b, \bm{x}_{0, j}^{\text{test}}}(\tfrac{iT}{N_T})\big)}{\sum_{j=1}^{N_{\text{test}}}\sum_{i=1}^{N_T} \E\Big[\big\|{b}(\bm{x}(\tfrac{iT}{N_T};\bm{x}_{0, j}^{\text{test}};\omega),\omega)\big\|^2\Big]},
\end{aligned}
    \label{sde_error}
\end{equation}
where $\bm{x}_{0, j}^{\text{test}}$ refers to the $j^{\text{test}}$ center in the testing set, $\mu^{\text{e}}_{\bm{x}_{0, j}^{\text{test}}}(\tfrac{iT}{N_T}), \hat{\mu}^{\text{e}}_{\bm{x}_{0, j}^{\text{test}}}(\tfrac{iT}{N_T})$ denotes the empirical distributions of the ground-truth and reconstructed trajectories at time $\tfrac{iT}{N_T}$ whose initial conditions are $\bm{x}_{0, j}^{\text{test}}$,
$\eta^{\text{e}}_{a, \bm{x}_{0, j}}$ and $\hat\eta^{\text{e}}_{a, \bm{x}_{0, j}}$ refers to the empirical distribution of $a(\bm{x}(t;\bm{x}_{0, j}^{\text{test}};\omega),\omega)$ and $\hat{a}(\bm{x}(t;\bm{x}_{0, j}^{\text{test}};\hat\omega),t;\hat{\omega})$, and  $\eta^{\text{e}}_{b, \bm{x}_{0, j}}$ and $\hat\eta^{\text{e}}_{b, \bm{x}_{0, j}}$ refers to the empirical distribution of $b(\bm{x}(t;\bm{x}_{0, j}^{\text{test}};\omega),\omega)$ and $\hat{b}(\bm{x}(t;\bm{x}_{0, j}^{\text{test}};\hat\omega),t;\hat{\omega})$, respectively.

 As shown in Table~\ref{tab:loss_compare}, 
using the Sinkhorn divergence is more computationally efficient, and the learned drift and diffusion functions are more accurate compared to those learned when using the temporally decoupled local squared $W_2$ as the loss function. The saving in runtime of using the temporal local Sinkhorn divergence Eq.~\eqref{temporal_avarage} is not significant compared to using the temporal local squared $W_2$ loss, and the possible reason is that the time needed to numerically solve the SDE is much longer than evaluating either the temporal local Sinkhorn divergence loss or the temporal local $W_2$ loss.

\begin{table}[ht]
\footnotesize
\centering
\caption{Comparison of the local Sinkhorn divergence versus the local squared $W_2$ loss to train an SNN for reconstructing the FHN model Eq.~\eqref{eq:fhn}.}
\label{tab:loss_compare}
\begin{tabular}{lcccc}
\toprule
Loss Function &
Training Time (s) &
Drift $W_2^2$ Error &
Diffusion $W_2^2$ Error&
Test trajectory $W_2^2$ Error \\
\midrule
Local Sinkhorn & 6772 & 0.1110 & 0.1090 & 0.06415\\
\midrule
Local squared $W_2$ & 7256 & 0.1195 & 0.1791 &0.06444\\
\bottomrule
\end{tabular}
\end{table}

\end{example}

\section{Summary and conclusion}
\label{section4}
In this paper, we proposed a local Sinkhorn divergence framework for conditional distribution reconstruction of random field models using SNNs. By utilizing the debiased Sinkhorn divergence, the proposed method preserved the geometric structure of optimal transport while introducing entropic regularization, leading to a differentiable and computationally efficient distribution matching objective. Compared with our previous local squared $W_2$ formulation, the proposed local Sinkhorn loss achieved improved computational efficiency and scalability while maintaining high reconstruction accuracy for learning multidimensional random field models. 
We further established theoretical error bounds for the proposed local Sinkhorn loss. The resulting bounds explicitly characterized the trade-off between approximation bias and statistical efficiency introduced by the entropic regularization parameter and implied that the proposed local Sinkhorn loss could partially alleviate the curse of dimensionality when learning multidimensional random field models by properly choosing the regularization parameter. 
Numerical experiments on one-dimensional conditional distribution reconstruction, stochastic Darcy flow, and stochastic FHN systems demonstrated that the proposed method accurately reconstructed random field models and outperformed several prevailing machine-learning uncertainty quantification benchmarks. Overall, the proposed local Sinkhorn approach
offered a practical balance between statistical accuracy, computational efficiency, and theoretical interpretability, making it a promising framework for uncertainty quantification and probabilistic scientific machine learning.

Several promising research directions deserve further investigation. First, the neighborhood radius $\delta$ and the regularization parameter $\varepsilon$ are currently selected empirically and remain fixed. Developing adaptive strategies for simultaneously choosing and adjusting these two parameters according to the local complexity of the conditional distribution may further improve both accuracy and computational efficiency. Recent advances in stabilized and multiscale Sinkhorn algorithms provide promising directions for adaptive regularization and efficient optimal transport computations \cite{schmitzer2019stabilized,feydy2019interpolating}.
Second, the current framework constructs neighborhoods using Euclidean distances. More general neighborhood constructions, including anisotropic neighborhoods, manifold-based neighborhoods, and graph-based local structures, may significantly improve the approximation of conditional distributions with low-dimensional intrinsic geometry \cite{belkin2003laplacian,tenenbaum2000global}. Such adaptive neighborhood constructions may also lead to improved theoretical convergence guarantees.
Another important direction is to extend the proposed local Sinkhorn framework to large-scale scientific machine learning problems, including stochastic partial differential equations, neural operator learning, probabilistic surrogate modeling, and Bayesian scientific machine learning. Since the proposed local Sinkhorn loss is fully differentiable and naturally compatible with GPU-based optimization, it is particularly suitable for these large-scale applications \cite{kovachki2023neuraloperator,raissi2019physics,papamakarios2021normalizing}.
Finally, it would be of considerable interest to further investigate the theoretical properties of local Sinkhorn learning in multidimensional settings. 

\section*{Declaration of generative AI and AI-assisted technologies in the manuscript preparation process} 
 During the preparation of this work, the authors used ChatGPT in order to polish the writing and assist in checking the manuscript. After using this tool/service, the authors reviewed and edited the content as needed and take full responsibility for the content of the published article.

\section*{Declaration of competing interest}
The authors declare that they have no known competing financial interests or personal relationships that could have
appeared to influence the work reported in this paper.

\section*{Data statement}

The data used in this study were generated numerically according to the models and experimental settings described in the manuscript. The data and implementation code that support the findings of this study will be made publicly available upon publication of the article.

\printcredits
\appendix

\section{Proof to Theorem~\ref{thm:sinkhorn_neighborhood_generalization}}
\label{app:neighborhood_sinkhorn_error}
Here, we prove Theorem~\ref{thm:sinkhorn_neighborhood_generalization}. We first decompose the generalization error as:
\begin{equation}
\begin{aligned}
\left|
\overline{\mathcal S}_{\varepsilon,\delta}^{\e}(\bm{y}_{\bm{x}},\hat{\bm{y}}_{\bm{x}})
-
\overline{\mathcal W}_{2}(\bm{y}_{\bm{x}},\hat{\bm{y}}_{\bm{x}})
\right|
\le\,
\left|
\overline{\mathcal S}_{\varepsilon,\delta}^{\e}(\bm{y}_{\bm{x}},\hat{\bm{y}}_{\bm{x}})
-
\overline{\mathcal W}_{2,\delta}^{\text{e}}(\bm{y}_{\bm{x}},\hat{\bm{y}}_{\bm{x}})
\right|
+
\left|
\overline{\mathcal W}_{2,\delta}^{\text{e}}(\bm{y}_{\bm{x}},\hat{\bm{y}}_{\bm{x}})
-
\overline{\mathcal W}_{2}^2(\bm{y}_{\bm{x}},\hat{\bm{y}}_{\bm{x}})
\right|,
\end{aligned}
\label{eq:triangle_sinkhorn_simple}
\end{equation}
where $\overline{\mathcal W}_{2}(\bm{y}_{\bm{x}},\hat{\bm{y}}_{\bm{x}})$ is defined in Eq.~\eqref{eq:globalsinkhorn}, and $\overline{\mathcal W}_{2,\delta}^{\text{e}}(\bm{y}_{\bm{x}},\hat{\bm{y}}_{\bm{x}})$ is the local squared $W_2$ distance between the two random fields $\bm{y}_{\bm{x}}$ and $\hat{\bm{y}}_{\bm{x}}$ defined as:
\begin{equation}
    \overline{\mathcal W}_{2,\delta}^{\text{e}}\big(\bm{y}_{\bm{x}},\hat{\bm{y}}_{\bm{x}}\big)\coloneqq  \int_{\mathcal{D}} W_2^2\big(\mu_{\bm{x}, \delta}^{\text{e}}, \hat{\mu}_{\bm{x}, \delta}^{\text{e}}\big)\nu(\d\bm{x}).
\end{equation}

The first term in Eq.~\eqref{eq:triangle_sinkhorn_simple} is the Sinkhorn regularization bias. By \cite[Theorem 1]{genevay2019sample} and the boundedness assumption (Assumption~\ref{assumptions}), taking the Lipschitz constant of the quadratic cost function $c$ in Eq.~\eqref{quadratic_cost} as $2M_0$ yields
\[
0\le W_\varepsilon(\alpha,\beta)-W(\alpha,\beta)
\le
2\varepsilon\log\!\left(
\frac{2e^2M_0^2}{\sqrt d\,\varepsilon}
\right),
\]
for every relevant pair of probability measures $(\alpha,\beta)$. Since the debiased Sinkhorn divergence is obtained by subtracting the self-transport terms, the same bound controls the difference between the local Sinkhorn loss and its $W_2^2$ counterpart, namely
\begin{equation}
\left|
\overline{\mathcal S}_{\varepsilon,\delta}^{\e}(\bm{y}_{\bm{x}},\hat{\bm{y}}_{\bm{x}})
-
\overline{\mathcal W}_{2,\delta}^{\e}(\bm{y}_{\bm{x}},\hat{\bm{y}}_{\bm{x}})
\right|
\le
2\varepsilon\log\!\left(
\frac{2e^2M_0^2}{\sqrt d\,\varepsilon}
\right).
\label{eq:sinkhorn_bias_bound}
\end{equation}

The second term in Eq.~\eqref{eq:triangle_sinkhorn_simple} is exactly the local squared $W_2$ generalization error. Therefore, applying \cite[Theorem~2.3]{xia2025efficient} in our previous work gives
\[
\mathbb E\!\left[
\left|
\overline{\mathcal W}_{2,\delta}^{\e}(\bm{y}_{\bm{x}},\hat{\bm{y}}_{\bm{x}})
-
\overline{\mathcal W}_{2}(\bm{y}_{\bm{x}},\hat{\bm{y}}_{\bm{x}})
\right|
\right]
\le
\frac{4M_0}{\sqrt{N}}
+
8CM_0\,\mathbb E\!\bigl[h(N(x,\delta),d)\bigr]
+
16M_0^2\,\delta.
\]

Combining the Sinkhorn bias bound in Eq.~\eqref{eq:sinkhorn_bias_bound} with the local $W_2$ generalization bound above proves Eq.~\eqref{eq:sinkhorn_gen_bound}.

\section{Proof to Theorem~\ref{thm:neighborhood_sinkhorn_error}}
\label{app:sinkhorn_neighborhood_generalization}
Here, we prove Theorem~\ref{thm:neighborhood_sinkhorn_error}.
We decompose the generalization error into three parts:
\begin{equation}
\begin{aligned}
\E\bigg[\left|
\overline{\mathcal S}_{\varepsilon,\delta}^{\e}(\bm{y}_{\bm{x}},\hat{\bm{y}}_{\bm{x}})
-
\overline{\mathcal W}_{2}^2(\bm{y}_{\bm{x}},\hat{\bm{y}}_{\bm{x}})
\right|\bigg]
\le\,
&
\E\bigg[\left|
\overline{\mathcal S}_{\varepsilon,\delta}^{\e}(\bm{y}_{\bm{x}},\hat{\bm{y}}_{\bm{x}})
-
\overline{\mathcal S}_{\varepsilon,\delta}(\bm{y}_{\bm{x}},\hat{\bm{y}}_{\bm{x}})
\right|\bigg]
\\
&+
\E\bigg[\left|
\int_D S_{\varepsilon}(\mu_{\bm{x}, \delta}, \hat{\mu}_{\bm{x}, \delta})\nu^{\text{e}}(\d\bm{x}) 
-
\int_D W_2^2(\mu_{\bm{x}, \delta}, \hat{\mu}_{\bm{x}, \delta})\nu^{\text{e}}(\d\bm{x})
\right|\bigg]\\
&+\E\bigg[\left|
\int_D W_2^2(\mu_{\bm{x}, \delta}, \hat{\mu}_{\bm{x}, \delta})\nu^{\text{e}}(\d\bm{x}) 
-
\int_D W_2^2(\mu_{\bm{x}}, \hat{\mu}_{\bm{x}})\nu^{\text{e}}(\d\bm{x})
\right|\bigg].
\end{aligned}
\label{eq:split_sinkhorn}
\end{equation}

The first term is the averaged empirical Sinkhorn error. For each $\bm{x}$, by applying \cite[Theorem~3]{genevay2019sample}, we have:
\begin{equation}
\mathbb E\!\left[
\left|
\mathcal S_{\varepsilon}(\mu_{\bm{x}, \delta}^{\text{e}},\hat \mu_{\bm{x}, \delta}^{\text{e}})
-
\mathcal S_{\varepsilon}(\mu_{\bm{x}, \delta},\hat \mu_{\bm{x}, \delta})
\right|
\right]
\le
\mathcal{O}
\Big(\frac{e^{\frac{5M_0^2}{\varepsilon}}}{\sqrt{N(\bm{x},\delta)}}
\Bigl(1+\varepsilon^{-\lfloor d/2\rfloor}\Bigr)\Big).
\label{eq:empirical_sinkhorn_bound}
\end{equation}

The second term is the average regularization bias between the Sinkhorn divergence and the Wasserstein discrepancy. From \cite[Theorem~1]{genevay2019sample}, for each $\bm{x}$, we have
\begin{equation}
\E\bigg[\left|
\mathcal S_{\varepsilon}(\mu_{\bm{x}, \delta},\hat \mu_{\bm{x}, \delta})
-
\mathcal W^2_{2}(\mu_{\bm{x}, \delta},\hat \mu_{\bm{x}, \delta})
\right|\bigg]
\le
2\varepsilon\log\!\left(
\frac{2e^2M_0^2}{\sqrt d\,\varepsilon}
\right).
\end{equation}
Therefore, 
\begin{equation}
    \E\bigg[\left|
\int_D S_{\varepsilon}(\mu_{\bm{x}, \delta}, \hat{\mu}_{\bm{x}, \delta})\nu^{\text{e}}(\d\bm{x}) 
-
\int_D W_2^2(\mu_{\bm{x}, \delta}, \hat{\mu}_{\bm{x}, \delta})\nu^{\text{e}}(\d\bm{x})
\right|\bigg]\le 2\varepsilon\log\!\left(
\frac{2e^2M_0^2}{\sqrt d\,\varepsilon}
\right).
\label{eq:sinkhorn_bias_bound1}
\end{equation}
Finally, for each $\bm{x}$, using the triangular inequality of the $W_2$ distance \cite{clement2008elementary}, we have:
\begin{equation}
    |W_2(\mu_{\bm{x}, \delta}, \hat{\mu}_{\bm{x}, \delta}) - W_2(\mu_{\bm{x}}, \hat{\mu}_{\bm{x}})|\leq W_2(\mu_{\bm{x}, \delta}, {\mu}_{\bm{x}})+ W_2(\hat\mu_{\bm{x}}, \hat{\mu}_{\bm{x}, \delta}).
\end{equation}

Following the proof of \cite[Theorem 2.3]{xia2026local} and the Lipschitz continuity condition in Assumption~\ref{assumptions}, we have:
\begin{equation}
    W_2(\mu_{\bm{x}, \delta}, {\mu}_{\bm{x}})+ W_2(\hat\mu_{\bm{x}}, \hat{\mu}_{\bm{x}, \delta})\leq 4M_0\delta.
    \label{neighbor_technique}
\end{equation}
Therefore,
\begin{equation}
\begin{aligned}
    |W_2^2(\mu_{\bm{x}, \delta}, \hat{\mu}_{\bm{x}, \delta}) - W_2^2(\mu_{\bm{x}}, \hat{\mu}_{\bm{x}})&|\leq |W_2(\mu_{\bm{x}, \delta}, \hat{\mu}_{\bm{x}, \delta}) - W_2(\mu_{\bm{x}}, \hat{\mu}_{\bm{x}})| \cdot |W_2(\mu_{\bm{x}, \delta}, \hat{\mu}_{\bm{x}, \delta}) + W_2(\mu_{\bm{x}}, \hat{\mu}_{\bm{x}})|\\
    &\leq 2(2M_0)\delta (2\sup\|y\|_2+2\sup\|\hat{y}\|_2) = 16M_0^2\delta,
    \end{aligned}
\end{equation}
which leads to:
\begin{equation}
    \E\Big[\big|W^2_{2}(\mu_{\bm{x}, \delta},\hat{\mu}_{\bm{x}, \delta})- W_{2}^2(\mu_{\bm{x}},\hat{\mu}_{\bm{x}})\big|\Big]\leq 16M_0^2\delta
    \label{L_w2}
\end{equation}

Combining Eqs.~\eqref{L_w2},  \eqref{eq:split_sinkhorn}, \eqref{eq:empirical_sinkhorn_bound}, and \eqref{eq:sinkhorn_bias_bound1} yields Eq.~\eqref{eq:neighborhood_sinkhorn_bound}. 

\textbf{Remark.} The error bound induced from using the neighborhood technique Eq.~\eqref{neighbor_technique} only applies to the Wasserstein distance through devising a special coupling of $\mu_{\bm{x}, \delta}$ and $\mu_{\bm{x}}$. However, there is no direct error bound for $S_{\varepsilon}(\mu_{\bm{x}, \delta}, \mu_{\bm{x}})$ as the Sinkhorn divergence introduces the additional entropic Kullback-Leibler divergence term. Therefore, it is nontrivial to obtain a direct error bound of 
\begin{equation}
    \left|
\overline{\mathcal S}_{\varepsilon,\delta}^{\e}(\bm{y}_{\bm{x}},\hat{\bm{y}}_{\bm{x}})
-
\int_{\mathcal{D}} S_{\varepsilon}(\mu_{\bm{x}}, \hat{\mu}_{\bm{x}}) \nu(\d\bm{x})
\right|,
\end{equation}
where $\int_{\mathcal{D}} S_{\varepsilon}(\mu_{\bm{x}}, \hat{\mu}_{\bm{x}}) \nu(\d\bm{x})$ denotes the average Sinkhorn divergence between $\bm{y}_{\bm{x}}$ and $\hat{\bm{y}}_{\bm{x}}$.

\section{Optimization, training settings and hyperparameters}
\label{app:hyper}

All numerical experiments are implemented in Python 3.11 using the PyTorch deep learning framework. The proposed Local Sinkhorn divergence is implemented using the GeomLoss package, while the $W_2$ distance is computed using the POT package. Unless otherwise specified, all competing methods within each numerical example employ the same neural network architecture, optimizer, initialization strategy, and training settings. All experiments are performed on a desktop equipped with an Intel Core i9-13900KF CPU (24 cores) and 64 GB RAM. The runtime reported in the numerical examples corresponds to the actual wall-clock training time under the same hardware environment. Training settings and hyperparameters are listed in Table~\ref{tab:hyperparameters}. In this work, the $W_2$ distance is evaluated numerically using the \texttt{ot.emd2} function in the \texttt{POT} package \cite{flamary2021pot} and the Sinkhorn divergence is evaluated numerically using the \texttt{SamplesLoss} function in the \texttt{Geomloss} package \cite{feydy2019interpolating}.






\begin{table*}[ht]
\centering
\caption{Training hyperparameters, neural network settings, and optimization configurations for all numerical examples.}
\label{tab:hyperparameters}
\renewcommand{\arraystretch}{1.2}
\begin{tabular}{lccc}
\toprule

&
Example \ref{example1}
&
Example \ref{example2}
&
Example \ref{example3}
\\

\midrule

Optimizer
&
Adam
&
Adam
&
Adam
\\

Learning rate
&
$5\times10^{-3}$
&
$5\times10^{-4}$
&
$1\times10^{-2}$
\\

Weight decay
&
0
&
0
&
0
\\

Number of epochs $E$
&
2000
&
20000
&
200
\\

Training samples
&
2000
&
2000
&
300
\\
Number of testing centers $N_c$
&
100
&
20
&
20
\\
Testing Realizations
&
20
&
100
&
20
\\
Mini-batch size $n_b$
&
8
&
4
&
8
\\

Hidden layers
&
2
&
3
&
2
\\

Hidden neurons
&
32
&
128
&
32
\\

Activation function
&
ReLU
&
ReLU
&
GeLU
\\

Residual connection (ResNet)
&
Yes
&
Yes
&
Yes
\\

Neighborhood radius $\delta$
&
0.05
&
0.05
&
0.1
\\
Minimal samples required for each neighborhood $n_{\min}$
&
4
&
4
&
4
\\
Maximal samples recorded in each neighborhood $n_{\max}$
&
128
&
64
&
300
\\
Samples chosen in each neighborhood $n$
&
32
&
32
&
32
\\

Regularization coefficient $\varepsilon$
&
0.05
&
0.03
&
0.1
\\

SNN parameter initialization
&
$\mathcal{N}(0, 0.01^2)$
&
$\mathcal{N}(0, 0.05^2)$
&
$\mathcal{N}(0, 0.01^2)$
\\
\bottomrule

\end{tabular}
\end{table*}


Unless otherwise specified, all model parameters are initialized using the default PyTorch initialization strategy. The same initialization scheme, optimizer, learning rate, training epochs, and network architecture are adopted for all competing methods within each numerical example to ensure a fair comparison.

\section{Definitions of Different Loss Metrics}
\label{app:loss}

Here, we summarize the loss functions used in Examples~\ref{example1} and \ref{example2}. Let
\[
B(\bm{x}_i,\delta)
=
\{\bm{x}_j:\|\bm{x}_j-\bm{x}_i\| \leq\delta\}
\]
denote the neighborhood centered at $\bm{x}_i$ (Eq.~\eqref{eq:neighbor}). If a center $\bm{x}_i$ is selected when evaluating the loss function within an epoch, a subset $\mathcal{B}(\bm{x}_i,\delta)\subseteq B(\bm{x}_i,\delta)$ is randomly chosen with $|\mathcal{B}(\bm{x}_i,\delta)| = \min(n, |B(\bm{x}_i,\delta)|)$. 
The overall loss is obtained by averaging the corresponding local loss over all randomly selected neighborhood centers.

\begin{itemize}
    \item 1. Local mean squared error (Local MSE)

\begin{equation}
\mathrm{MSE}_\delta(\bm{y}_{\bm{x}},\hat{\bm{y}}_{\bm{x}})
=
\frac1{n_b}
\sum_{i=1}^{n_b}
\frac1{|\mathcal B(\bm{x}_i,\delta)|}
\sum_{\bm{x}_j\in\mathcal B(\bm{x}_i,\delta)}
\|y(\bm{x}_j;\omega_j)-\hat{\bm{y}}(\bm{x}_j;\hat\omega_j)\|^2 .
\end{equation}

\item 2. Local mean absolute error (Local MAE)

\begin{equation}
\mathrm{MAE}_\delta(\bm{y}_{\bm{x}},\hat{\bm{y}}_{\bm{x}})
=
\frac1{n_b}
\sum_{i=1}^{n_b}
\frac1{|\mathcal B(\bm{x}_i,\delta)|}
\sum_{\bm{x}_j\in\mathcal B(\bm{x}_i,\delta)}
\|
\bm y(\bm{x}_j;\omega_j)
-
\hat{\bm y}(\bm{x}_j;\hat\omega_j)
\|.
\end{equation}

\item 3. Local energy distance (Local ED)

The energy distance follows from \cite{szekely2013energy} and is defined locally by

\begin{equation}
\mathrm{ED}_\delta(\bm{y}_{\bm{x}},\hat{\bm{y}}_{\bm{x}})
=
\frac1{n_b}
\sum_{i=1}^{n_b}
\mathrm{ED}
\!\left(
Y_i^\delta,
\hat Y_i^\delta
\right),
\end{equation}

where

\begin{equation}
\begin{aligned}
\mathrm{ED}
\!\left(
Y_i^\delta,
\hat Y_i^\delta
\right)
&=\frac{2}{|\mathcal B(\bm{x}_i,\delta)|^2}
\sum_{\bm{y}_p\in Y_i^{\delta}, \hat{\bm{y}}_q\in \hat{Y}_i^{\delta}}
\|
\bm y_p
-
\hat{\bm y}_q
\|
\\&\qquad\qquad-\frac1{|\mathcal B(\bm{x}_i,\delta)|^2}
\sum_{\bm{y}_p\in Y_i^{\delta}, \hat{\bm{y}}_q\in \hat{Y}_i^{\delta}}
\|
\bm y_p
-
\bm y_q
\|
-\frac1{|\mathcal B(\bm{x}_i,\delta)|^2}
\sum_{\bm{y}_p\in Y_i^{\delta}, \hat{\bm{y}}_q\in \hat{Y}_i^{\delta}}
\|
\hat{\bm y}_p
-
\hat{\bm y}_q
\|,
\end{aligned}
\end{equation}
with
\begin{equation}
Y_i^\delta
=
\left\{
\bm y(\bm{x}_j;\omega_j):
\bm{x}_j\in\mathcal B(\bm{x}_i,\delta)
\right\},
\qquad
\hat Y_i^\delta
=
\left\{
\hat{\bm y}(\bm{x}_j;\hat\omega_j):
\bm{x}_j\in\mathcal B(\bm{x}_i,\delta)
\right\}.
\label{Y_def}
\end{equation}

\item 4. Local maximum mean discrepancy (Local MMD)

The MMD follows from the kernel two-sample test \cite{gretton2012kernel} and is defined locally by:
\begin{equation}
\mathrm{MMD}_\delta(\bm{y}_{\bm{x}},\hat{\bm{y}}_{\bm{x}})
=
\frac{1}{|\Gamma|}\sum_{\gamma\in \Gamma}\frac1{n_b}
\sum_{i=1}^{n_b}
\mathrm{MMD}_{\gamma}
\!\left(
Y_i^\delta,
\hat Y_i^\delta
\right),
\end{equation}
where
\begin{equation}
\begin{aligned}
\mathrm{MMD}_{\gamma}
\!\left(
Y_i^\delta,
\hat Y_i^\delta
\right)
&=\frac1{|\mathcal B(\bm{x}_i,\delta)|^2}
\sum_{\bm{y}_p\in Y_i^{\delta}, \hat{\bm{y}}_q\in \hat{Y}_i^{\delta}}
K_{\gamma}(\bm y_p,\bm y_q)
+
\frac1{|\mathcal B(\bm{x}_i,\delta)|^2}
\sum_{\bm{y}_p\in Y_i^{\delta}, \hat{\bm{y}}_q\in \hat{Y}_i^{\delta}}
K_{\gamma}(\hat{\bm y}_p,\hat{\bm y}_q)
\\&\qquad\qquad-
\frac2{|\mathcal B(\bm{x}_i,\delta)|^2}
\sum_{\bm{y}_p\in Y_i^{\delta}, \hat{\bm{y}}_q\in \hat{Y}_i^{\delta}}
K_{\gamma}(\bm y_p,\hat{\bm y}_q),
\end{aligned}
\end{equation}
where $Y_i^{\delta}$ and $\hat{Y}_i^{\delta}$ are defined in Eq.~\eqref{Y_def}, and \(K(\cdot,\cdot)\) denotes the Gaussian kernel:
\[
K_{\gamma}(\bm y,\hat{\bm y})
=
\exp\!\left(
-\frac{\|\bm y-\hat{\bm y}\|^2}{2\gamma^2}
\right).
\]
In Examples~\ref{example1} and \ref{example2}, $\Gamma=\{0.5, 1.0, 2.0, 4.0\}$

\item 5. Local squared $W_2$ distance

\begin{equation}
W_{2,\delta}^2(\bm{y}_{\bm{x}},\hat{\bm{y}}_{\bm{x}})
=
\frac1{n_b}
\sum_{i=1}^{n_b}
W_2^2
(\mu_{\bm{x}_i, \delta}^{\text{e}},\hat\mu_{\bm{x}_i, \delta}^{\text{e}}),
\end{equation}
where $\mu_{\bm{x}_i, \delta}^{\text{e}}$ and $\hat\mu_{\bm{x}_i, \delta}^{\text{e}}$ are the empirical distributions associated with the samples $\bm{y}(\bm{x}_j;\omega_j)$ in $Y_i^{\delta}$ and $\hat{\bm{y}}(\bm{x}_j;\hat{\omega}_j)$ in $\hat{Y}_i^{\delta}$ defined in Eq.~\eqref{Y_def}, respectively. 


\item 6. Local Sinkhorn divergence

This loss is adopted throughout the numerical experiments unless otherwise specified.
\begin{equation}
S_{\varepsilon,\delta}
(\bm{y}_{\bm{x}}, \hat{\bm{y}}_{\bm{x}})
=
\frac1{n_b}
\sum_{i=1}^{n_b}
S_\varepsilon
(\mu_{\bm{x}_i, \delta}^{\text{e}},\hat\mu_{\bm{x}_i, \delta}^{\text{e}}),
\end{equation}
where $S_{\varepsilon}(\mu, \hat{\mu})$ is the Sinkhorn divergence defined in Eq.~\eqref{eq:sinkhorndiv}, and 
$\mu_{\bm{x}_i, \delta}^{\text{e}}$ and $\hat\mu_{\bm{x}_i, \delta}^{\text{e}}$ are the empirical distributions associated with the samples $\bm{y}(\bm{x}_j;\omega_j)$ in $Y_i^{\delta}$ and $\hat{\bm{y}}(\bm{x}_j;\hat{\omega}_j)$ in $\hat{Y}_i^{\delta}$ defined in Eq.~\eqref{Y_def}, respectively.
\end{itemize}

\bibliographystyle{plain}
\bibliography{bibliography}

\end{document}